\documentclass[acmtog,nonacm]{acmart}
\acmSubmissionID{310}

\newcommand{\opa}{\sigma}

\usepackage{booktabs}
\usepackage{xspace}

\usepackage[ruled]{algorithm2e}

\SetAlFnt{\small}
\SetAlCapFnt{\small}
\SetAlCapNameFnt{\small}
\SetAlCapHSkip{0pt}

\usepackage[capitalise,nameinlink]{cleveref}
\Crefname{equation}{Eq.}{Eqs.}
\Crefname{figure}{Fig.}{Figs.}
\Crefname{tabular}{Tab.}{Tabs.}
\Crefname{section}{Sec.}{Secs.}

\usepackage{mathtools}
\usepackage{makecell}
\usepackage{pifont}
\definecolor{CheckGreen}{rgb}{0, 0.55, 0}
\definecolor{XRed}{RGB}{180,0,0}
\newcommand{\xmark}{{\color{XRed}\ding{55}}}

\usepackage[table]{xcolor}
\usepackage{colortbl}
\definecolor{yellow}{rgb}{1, 1, 0.7}
\definecolor{orange}{rgb}{1, 0.85, 0.7}
\definecolor{red}{rgb}{1, 0.7, 0.7}

\acmJournal{TOG}

\newcommand{\method}{\textsc{ArtiFixer}\xspace}
\newcommand{\methodTD}{\textsc{ArtiFixer3D}\xspace}
\newcommand{\methodTDp}{\textsc{ArtiFixer3D+}\xspace}

\begin{document}
\title{\method: Enhancing and Extending 3D Reconstruction with Auto-Regressive Diffusion Models}

\author{Riccardo de Lutio}
\orcid{0000-0002-2644-3876}
\authornote{Equal contribution.}
\affiliation{%
  \institution{NVIDIA}
   \city{Santa Clara}
   \country{USA}
  }
\email{rdelutio@nvidia.com}

\author{Tobias Fischer}
\orcid{0000-0001-8227-001X}
\affiliation{%
  \institution{NVIDIA}
   \city{Zurich}
   \country{Switzerland}
}
\affiliation{%
  \institution{ETHZ}
   \city{Zurich}
   \country{Switzerland}
}
\email{tobiasf@nvidia.com}

\author{Yen-Yu Chang}
\orcid{0009-0006-4105-1770}
\affiliation{%
  \institution{NVIDIA}
   \city{Santa Clara}
   \country{USA}
}
\affiliation{%
  \institution{Cornell University}
   \city{Ithaca}
   \country{USA}
}
\email{yc2463@cornell.edu}

\author{Yuxuan Zhang}
\orcid{0000-0002-6409-5550}
\affiliation{%
  \institution{NVIDIA}
   \city{New York}
   \country{USA}
}
\email{alezhang@nvidia.com}

\author{Jay Zhangjie Wu}
\orcid{0009-0003-3684-7262}
\affiliation{%
  \institution{NVIDIA}
   \city{Toronto}
   \country{Canada}
}
\email{wjay@nvidia.com}

\author{Xuanchi Ren}
\orcid{0000-0001-6376-7100}
\affiliation{%
  \institution{NVIDIA}
   \city{Toronto}
   \country{Canada}
}
\affiliation{%
  \institution{University of Toronto}
   \city{Toronto}
   \country{Canada}
}
\affiliation{%
  \institution{Vector Institute}
   \city{Toronto}
   \country{Canada}
}
\email{xuanchir@nvidia.com}

\author{Tianchang Shen}
\orcid{0000-0002-7133-2761}
\affiliation{%
  \institution{NVIDIA}
   \city{Toronto}
   \country{Canada}
}
\affiliation{%
  \institution{University of Toronto}
   \city{Toronto}
   \country{Canada}
}
\affiliation{%
  \institution{Vector Institute}
   \city{Toronto}
   \country{Canada}
}
\email{frshen@nvidia.com}

\author{Katar\'{\i}na T\'{o}thov\'{a}}
\orcid{0000-0001-5864-179X}
\affiliation{%
  \institution{NVIDIA}
   \city{Zurich}
   \country{Switzerland}
}
\email{ktothova@nvidia.com}

\author{Zan Gojcic}
\orcid{0000-0001-6392-2158}
\affiliation{%
  \institution{NVIDIA}
   \city{Zurich}
   \country{Switzerland}
}
\email{zgojcic@nvidia.com}

\author{Haithem Turki}
\authornotemark[1]
\orcid{0000-0001-5634-0918}
\affiliation{%
  \institution{NVIDIA}
   \city{Seattle}
   \country{USA}
}
\email{hturki@nvidia.com}

\renewcommand\shortauthors{de Lutio, R. et al}

\begin{abstract}
Per-scene optimization methods such as 3D Gaussian Splatting provide state-of-the-art novel view synthesis quality but extrapolate poorly to under-observed areas. Methods that leverage generative priors to correct artifacts in these areas hold promise but currently suffer from two shortcomings. The first is scalability, as existing methods use image diffusion models or bidirectional video models that are limited in the number of views they can generate in a single pass (and thus require a costly iterative distillation process for consistency). The second is quality itself, as generators used in prior work tend to produce outputs that are inconsistent with existing scene content and fail entirely in completely unobserved regions. To solve these, we propose a two-stage pipeline that leverages two key insights. First, we train a powerful bidirectional generative model with a novel opacity mixing strategy that encourages consistency with existing observations while retaining the model's ability to extrapolate novel content in unseen areas. Second, we distill it into a causal auto-regressive model that generates hundreds of frames in a single pass. This model can directly produce novel views or serve as pseudo-supervision to improve the underlying 3D representation in a simple and highly efficient manner. We evaluate our method extensively and demonstrate that it can generate plausible reconstructions in scenarios where existing approaches fail completely. When measured on commonly benchmarked datasets, we outperform all existing baselines by a wide margin, exceeding prior state-of-the-art methods by 1-3 dB PSNR.
\end{abstract}

\begin{teaserfigure}
\centering
\setlength{\abovecaptionskip}{2pt}
\includegraphics[width=1\textwidth]{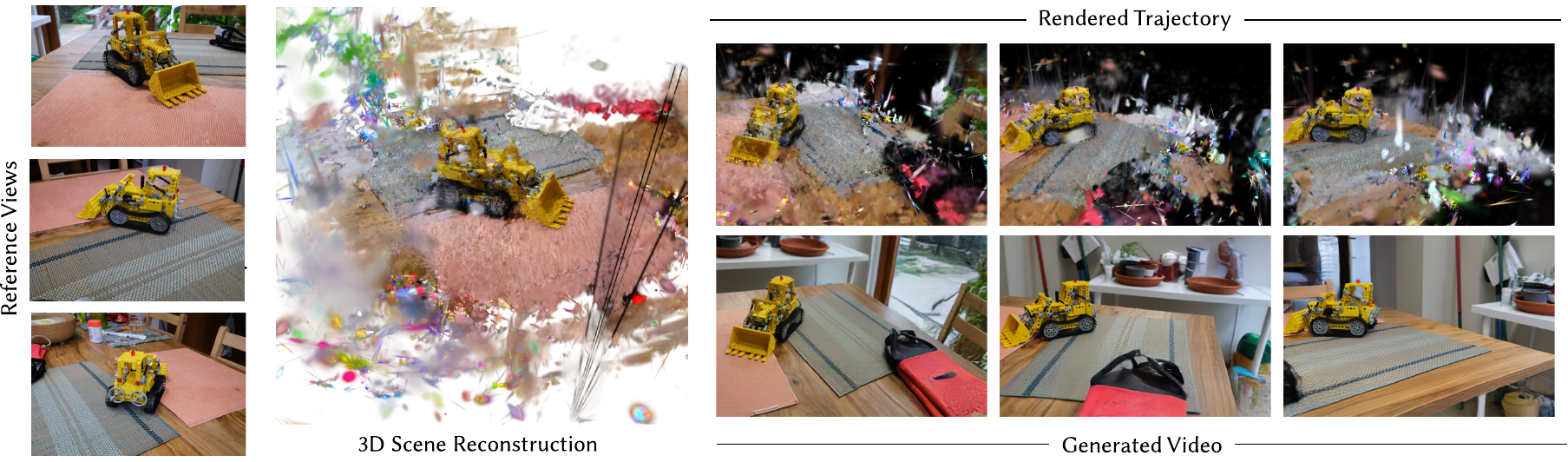}
\caption{\method enhances and extends existing 3D reconstructions in a highly efficient and scalable manner. Given an initial reconstruction and optional reference views and text prompt, it auto-regressively generates novel content that maintains a high degree of consistency with existing observations. \method can directly produce hundreds of novel views in a single inference pass or serve as pseudo-supervision to improve the underlying 3D reconstruction.
\bfseries Project page: \href{https://research.nvidia.com/labs/sil/projects/artifixer}{https://research.nvidia.com/labs/sil/projects/artifixer}\label{fig:teaser}}
\end{teaserfigure}


\maketitle

\section{Introduction}
\label{sec:intro}

High-quality novel view synthesis is essential for applications in virtual and augmented reality and closed-loop simulation for physical AI. These use cases require photorealistic rendering and the ability to navigate complex environments under unconstrained camera motion. In recent years, two paradigms have emerged as dominant approaches to novel view synthesis: explicit 3D neural reconstruction~\cite{mildenhall2020nerf, kerbl3Dgaussians}, and camera-controlled image or video generation~\cite{ren2025gen3c, zhou2025stable}.
 
Neural reconstruction methods have matured significantly and now enable real-time rendering and high visual fidelity when trained from dense image collections with accurate camera poses. However, in the most widely used per-scene optimization setting, their performance remains fundamentally limited by the completeness and quality of the input observations. Regions that are sparsely observed or entirely missing during capture are poorly reconstructed, leading to artifacts, holes, or implausible geometry. While such deficiencies remain hidden near the training views, they are inevitably exposed during free navigation of the scene.

Conversely, recent video generative models have demonstrated the ability to synthesize photorealistic and temporally coherent content that is often indistinguishable from real-world videos~\cite{veo2024, sora2024, nvidia2025cosmosworldfoundationmodel}. Despite this progress, precise camera control over extended sequences, long-term temporal consistency, and the accumulation of drift and hallucinations remain open challenges, limiting their applicability to interactive view synthesis.

Instead of treating reconstruction and generation as standalone alternatives, we aim to combine their complementary strengths: generative models serve as powerful priors to repair and complete imperfect reconstructions, while the explicit—albeit noisy and partial—3D representation provides a strong conditioning signal that grounds generation, mitigates long-term drift, and suppresses hallucinations. Recent methods have taken initial steps in this direction by training generative models to map degraded novel-view renderings to clean images and distilling the resulting improvements back into an underlying 3D representation~\cite{gao2024cat3d, yu2024viewcrafter, wu2025difix3d+, fischer2025flowr}. However, these approaches must navigate two fundamental trade-offs. First, they must balance temporal consistency and efficiency: some employ large bidirectional video generative models that provide strong temporal coherence but incur high computational cost~\cite{gao2024cat3d, Wu2025GenFusion, fischer2025flowr}, while others rely on (multi-view) image-based generative models that are more efficient but limit temporal consistency and require progressive distillation strategies~\cite{Rundi2024CVPR, wu2025difix3d+}. Second, they face the trade-off between conditioning strength and generative capacity. Approaches~\cite{yu2024viewcrafter, Wu2025GenFusion} that condition generation on corrupted renderings via concatenation or cross-attention risk altering the observed scene content, whereas methods~\cite{wu2025difix3d+, fischer2025flowr} trained to directly map corrupted renderings to clean images are incapable of synthesizing missing content, due to the mode collapse in fully unobserved regions where all input pixels are black.

In our work, we follow this line of research by adapting a pretrained bidirectional video diffusion model into a camera-controllable generator that maps corrupted renderings to clean images. To overcome the aforementioned limitations, we introduce two key contributions: \textbf{(i)} an opacity-aware noise mixing strategy that injects Gaussian noise into low-opacity regions, preventing mode collapse and preserving generative capacity in unobserved areas; and \textbf{(ii)} distillation of the bidirectional model into a few-step causal auto-regressive generator capable of producing arbitrarily long, temporally consistent videos while approaching the efficiency of prior image-based methods. In doing so, we demonstrate that even highly degraded 3D reconstructions provide sufficient conditioning signals to significantly simplify the distillation process. While recent work has begun incorporating explicit 3D representations as conditioning signals for auto-regressive video generation~\cite{zhai2025stargen, wu2025spmem, he2025flexworld}, these approaches treat the 3D input as a fixed conditioning rather than an output to be improved. Our method closes this loop: the reconstruction conditions the generator, and the generator in turn enhances and extends the reconstruction, enabling both higher-quality video synthesis and improved 3D scene completeness. The resulting framework enables efficient improvement of the underlying 3D reconstruction and greatly outperforms a wide range of baselines across multiple benchmarks.

\section{Related Work}
\label{sec:related}

\paragraph{Novel view synthesis from 3D representations.} Neural Radiance Fields (NeRFs)~\cite{mildenhall2020nerf} and, more recently, 3D Gaussian Splatting (3DGS)~\cite{kerbl3Dgaussians} have revolutionized the field of novel view synthesis by distilling sensor information (usually overlapping photos of a scene) into a 3D representation that can then be queried from arbitrary camera viewpoints. Because these representations are optimized on a per-scene basis, their ability to extrapolate beyond observed views is inherently limited, and they fail to render plausible content in sparsely observed or missing regions.

A large body of work seeks to mitigate these limitations through handcrafted geometric priors~\cite{Niemeyer2021Regnerf, Yang2023FreeNeRF, somraj2023simplenerf}, pretrained depth~\cite{deng2022depth, roessle2022dense, wang2023sparsenerf, zhu2023FSGS} and normal~\cite{yu2022monosdf} estimators, and adversarial networks~\cite{roessle2023ganerf}. However, these approaches are sensitive to noise, difficult to balance with data terms, and yield only marginal improvements in denser captures. An alternative line of work trains feed-forward networks on large multi-scene datasets, which are used to enhance a scene-optimized NeRF/3DGS~\cite{zhou2023nerflix, lu2025matrix3d} or directly predict novel views~\cite{yu2020pixelnerf, mvsnerf, ren2024scube, lu2025infinicube}. While these deterministic methods perform well near reference views, they often produce blurry results in ambiguous regions where the distribution of possible renderings is inherently multi-modal.

\begin{figure*}
  \centering
    \includegraphics[width=\textwidth]{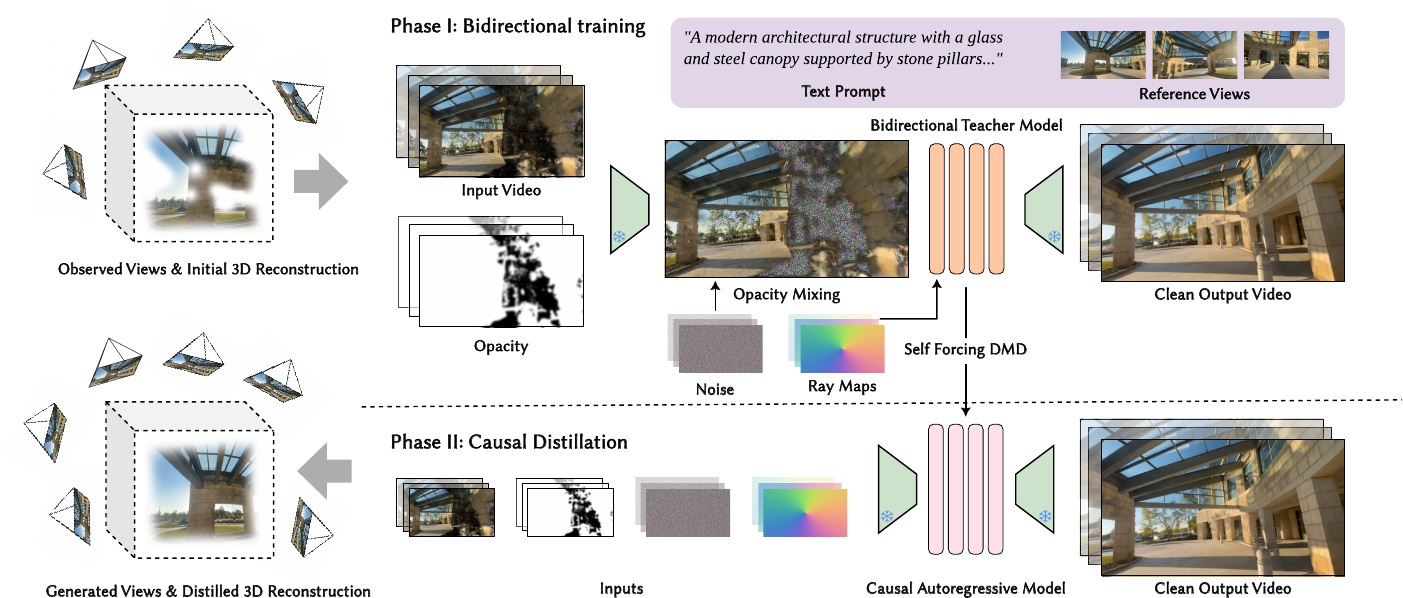}
    \caption{\textbf{Method overview.} We first train a bidirectional flow matching model that transports degraded RGB renderings into clean outputs. We encode the input RGB into latent space and mix with Gaussian noise using the rendered opacity maps to avoid mode collapse in unseen regions. We inject fine-grained opacity information and camera control along with optional clean reference views and a text prompt. In the second phase of our pipeline, we distill the teacher into an auto-regressive causal model via Self Forcing-style DMD distillation~\cite{huang2025selfforcing}, which can be directly used to render novel views or used as pseudo-supervision to distill back into the underlying 3D representation.} 
    \label{fig:method-overview}
\end{figure*}

\paragraph{Diffusion models for novel view synthesis.} An alternative strategy is to leverage the priors learned by generative diffusion models trained on internet-scale data to enhance novel view synthesis. Early works~\cite{poole2023dreamfusion, Kyle2024CVPR, Rundi2024CVPR} use a diffusion model as a learned critic during reconstruction optimization, but this incurs substantial computational overhead. More recent approaches~\cite{gao2024cat3d, liu20243dgs, liu2023deceptive, wu2025difix3d+, Wu2025GenFusion, fischer2025flowr} directly generate multi-view–consistent images that can be consumed by a downstream 3D reconstruction pipeline. While this strategy improves training efficiency, it typically relies on iterative generation and distillation, in which new views are progressively distilled back into the 3D representation to satisfy computational and consistency constraints. Lyra~\cite{bahmani2026lyra} sidesteps this iteration by distilling video diffusion knowledge into a feed-forward 3DGS generator, though it operates from a single image rather than enhancing an existing reconstruction.
Recent work reverses this paradigm by building on the rapid progress of video generation~\cite{blattmann2023stable, wan2025}. Rather than distilling generative outputs into a 3D representation, these methods treat the 3D representation as a conditioning signal for a generative model that directly synthesizes novel views~\cite{ren2025gen3c, kong2025worldwarp}. Although this approach can improve the perceptual realism of novel views, it inherits limitations of the underlying generative models, including temporal inconsistencies, hallucinations, and imperfect camera control.

\paragraph{Auto-regressive video generation.} While bidirectional video generation models synthesize all frames jointly, auto-regressive models generate frames sequentially using block-causal attention. Auto-regressive generation improves scalability and generation efficiency compared to bidirectional models, but often suffers from quality degradation over time, as each frame is conditioned on previously generated outputs, causing errors to accumulate~\cite{yin2025causvid}. Several methods try to address the issue by better aligning the training scheme of these models with inference-time conditions, thereby reducing exposure bias~\cite{huang2025selfforcing, cui2025self, liu2025rolling}. A complementary line of research focuses on improving generation speed and controllability by exploiting temporal and spatial cues to select per-frame context~\cite{yang2025longlive, kong2025worldwarp, shin2025motionstream, huang2025voyager, li2025vmem}, enabling interactive auto-regressive world models~\cite{hong2025relicinteractivevideoworld}.
Despite these advances, auto-regressive video models still lag behind explicit 3D representations in terms of spatial consistency, camera controllability, and rendering efficiency.

\section{Preliminaries}
\label{sec:background}

\paragraph{3D Gaussian Splatting.} 3DGS~\cite{kerbl3Dgaussians} represents a scene as a set of anisotropic 3D Gaussian primitives, each parameterized by a mean $\boldsymbol{\mu}_j$, covariance $\boldsymbol{\Sigma}_j$, opacity $\opa_j$, and view-dependent color $\mathbf{c}_j$. Novel views are rendered by projecting the primitives onto the target image plane and compositing in front-to-back depth order: $\mathcal{C}(\mathbf{p}) = \sum_{i} \alpha_i \mathbf{c}_i \prod_{k<i}(1-\alpha_k)$, where $\alpha_i$ is the learned opacity scaled by the projected Gaussian evaluated at pixel $\mathbf{p}$. Primitive parameters are optimized per scene with a photometric reconstruction loss.

\paragraph{Video diffusion models.}
Diffusion models learn to transport samples between a data distribution $p_{data}(\mathbf{x})$ and a tractable prior, typically $\mathcal{N}(\mathbf{0},\mathbf{I})$~\cite{song2020score, ho2020diffusionmodels}. Most video diffusion models~\cite{blattmann2023stable} operate in a lower-dimensional latent space for computational efficiency. Flow matching~\cite{lipman2023flowmatching, liu2023flow}, the framework used by our method, learns an ODE flow between two arbitrary endpoint distributions $p_{src}$ and $p_{tgt}$ by fitting a time-dependent vector field $\mathbf{v}_\theta(\mathbf{z}_t,t)$ whose induced probability path $\{p_t\}_{t\in[0,1]}$ satisfies $p_0=p_{src}$ and $p_1=p_{tgt}$. During training, we sample endpoint latents $\mathbf{z}_0\sim p_{src}$ and $\mathbf{z}_1\sim p_{tgt}$ and a time $t\in[0,1]$, construct an intermediate latent via $\mathbf{z}_t \coloneqq (1-t)\mathbf{z}_0+t\mathbf{z}_1$ with target velocity $\mathbf{v}_t \coloneqq \frac{d\mathbf{z}_t}{dt}=\mathbf{z}_1-\mathbf{z}_0$, and fit the vector field using the conditional flow matching objective $\min_{\theta}\ \mathbb{E}_{t,\mathbf{z}_0,\mathbf{z}_1}\bigl\lVert \mathbf{v}_\theta(\mathbf{z}_t,t)-\mathbf{v}_t\bigr\rVert_2^2$. At inference, we draw $\mathbf{z}_0\sim p_{src}$ and numerically integrate the learned ODE from $t=0$ to $t=1$ to obtain $\mathbf{z}_1$ as a sample from $p_{tgt}$.

\section{Method}
\label{sec:method}

Given an initial 3D reconstruction of a scene created from a sparse set of images, our goal is to generate artifact-free renderings from arbitrary camera viewpoints, including regions unobserved by input images, at interactive rates.
Our solution is a controllable auto-regressive video model that can either directly render arbitrary long novel-view renderings or provide pseudo-supervision to improve the underlying 3D reconstruction. We describe how to adapt a pretrained video diffusion model to serve as a bidirectional teacher in \cref{sec:bidirectional-training}. We discuss causal distillation and the capabilities of the resulting model in \cref{sec:causal-distillation}. \cref{fig:method-overview} illustrates our approach.

\subsection{Bidirectional Training}
\label{sec:bidirectional-training}

\begin{figure}
  \centering
    \includegraphics[width=\linewidth]{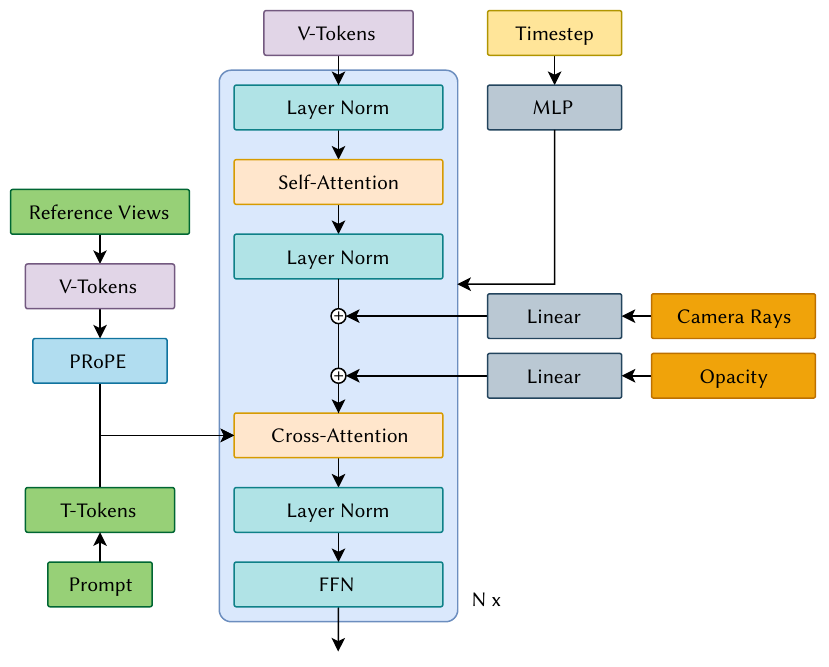}
    \caption{\textbf{Transformer block.} We start from a pretrained text-to-video model~\cite{wan2025} and inject camera and opacity information into each transformer block via linear layers after applying self-attention and layer normalization. We patchify reference views into visual tokens, apply relative camera conditioning via PRoPE~\cite{li2025cameras}, and add $K_n$ and $V_n$ projections to the cross-attention operation. We zero-initialize $f_r$, $f_o$, and $V_n$ to ensure compatibility with the pretrained initialization.}
    \label{fig:transformer-block}
\end{figure}

\paragraph{Architecture.}
We start from a pretrained text-to-video model (Wan 2.1 T2V-14B~\cite{wan2025}), freeze its VAE and text encoder, and finetune the remaining components. Degraded renderings are encoded by the frozen VAE and 3D-patchified with $(t,h,w)=(1,2,2)$, where $(t,h,w)$ is the temporal/vertical/horizontal patch size in latent voxels. We guide where to generate scene content through rendered opacity maps $\mathbf{O}$ and enable camera control in completely unobserved areas via per-pixel Plücker raymaps $\mathbf{R}$, which assign each pixel the six-vector $(\mathbf{d},\,\mathbf{o}\times\mathbf{d})$ formed from its ray direction $\mathbf{d}$ (unprojected through the camera intrinsics/extrinsics) and the camera center $\mathbf{o}$.
Both signals bypass the VAE entirely -- we downscale their spatial dimensions to match the spatial compression factor of the VAE via the PixelUnshuffle operation~\cite{paszke2019pytorch}, encode them via per-block linear layers $f_o$ and $f_r$ (\cref{fig:transformer-block}), and add the embeddings to the visual tokens:
\begin{align}
T_r := T_s + f_r(\text{PixelUnshuffle}(\mathbf{R})) \\
   T_o := T_r + f_o(\text{PixelUnshuffle}(\mathbf{O})), \label{eq:fine-opacity}
\end{align}
where $T_s$ denotes the token set after applying self-attention and layer-normalization. We found this strategy to be more computationally efficient than alternatives such as VAE encoding $\mathbf{R}$ and $\mathbf{O}$ while providing camera control even when the input rendering is entirely empty. To provide additional scene context, we encode clean reference views with the frozen VAE, patchified per-image along the batch dimension (no temporal compression). Each transformer block then cross-attends from target tokens ($Q$) to the concatenated reference tokens, which are mapped to keys and values via additional linear projections $K_n$ and $V_n$; the cross-attention output is added back to the target tokens, following the image-to-video variant of Wan 2.1. We apply PRoPE~\cite{li2025cameras} only within this cross-attention, using target intrinsics/extrinsics for $Q$ and reference intrinsics/extrinsics for $K_n/V_n$. $f_r$, $f_o$, and $V_n$ are all zero-initialized to ensure compatibility with the pretrained initialization.

\begin{figure*}
  \centering
    \includegraphics[width=\textwidth]{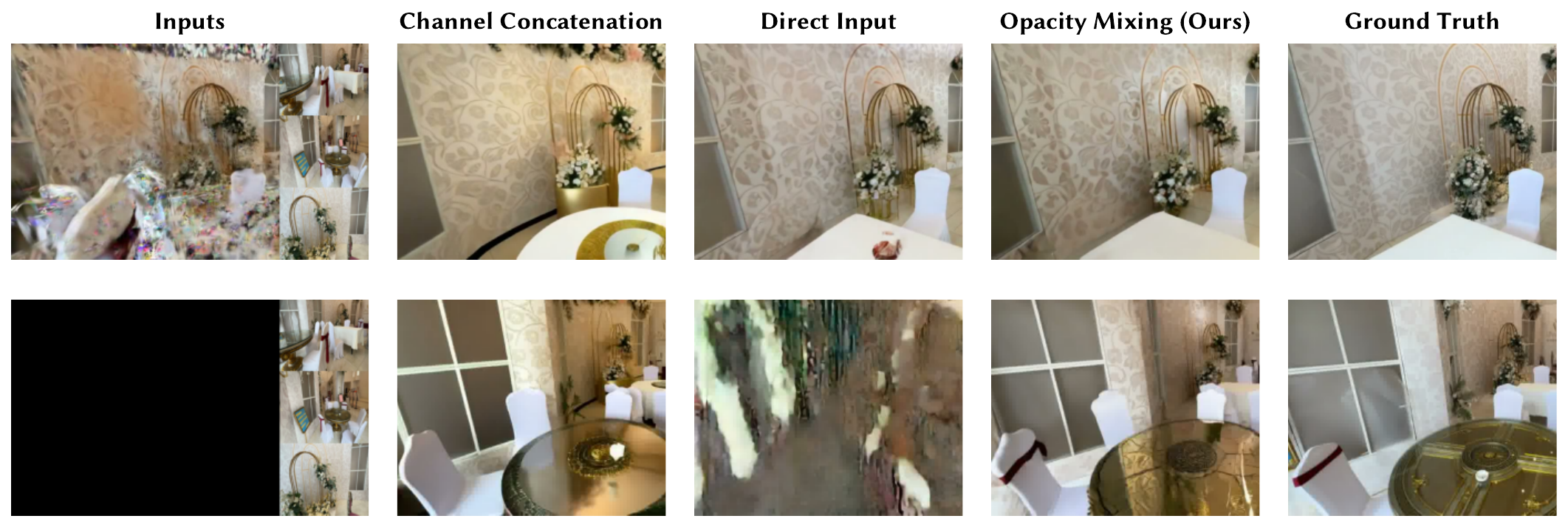}
    \caption{\textbf{Opacity mixing.} Given a degraded rendering and optional reference views and text prompt (\textbf{left}), we predict an artifact-free rendering at a target viewpoint. Starting from Gaussian noise and channel concatenating the degraded rendering as in prior work~\cite{Wu2025GenFusion, yin2025gsfixerimproving3dgaussian} produces renderings that are semantically similar to the reference views, but with notable inconsistencies (such as the table in the~\textbf{top row}). Directly starting from the degraded rendering instead of Gaussian noise improves consistency, but degrades quality noticeably when extrapolating to areas outside those covered by the degraded renderings (\textbf{bottom row}). Instead, we mix Gaussian noise into the rendering based on its opacity map. The resulting input retains the consistency benefits of the original while enabling a strong generative capability in entirely novel regions.}
    \label{fig:opacity-mixing}
\end{figure*}

\paragraph{Opacity mixing.} Most generative models start from Gaussian noise $\boldsymbol{\epsilon} \sim \mathcal{N}(\mathbf{0}, \mathbf{I})$ which is iteratively transformed into a latent video representation $\mathbf{z}$. Most prior work similarly starts from such noise, conditioning the generation process on the initial degraded rendering latent $\mathbf{z}_{deg}$ via channel-concatenation~\cite{Wu2025GenFusion, yin2025gsfixerimproving3dgaussian} or classifier-free guidance~\cite{liu20243dgs}. Although the resulting latent $\mathbf{z}_{enh}$ tends to be semantically similar to its degraded counterpart, notable inconsistencies remain, especially in high-artifact regions (\cref{fig:opacity-mixing}). Several methods start directly from $\mathbf{z}_{deg}$ instead of noise~\cite{wu2025difix3d+, fischer2025flowr}, validating the insight that the source distribution should reflect what can already be rendered. While this encourages stronger consistency guarantees, it suffers from mode collapse in completely unseen areas: the source distribution collapses to a Dirac mass at zero in empty regions, hindering the ability to extrapolate high-quality renderings (\cref{fig:opacity-mixing}). To address this, we mix Gaussian noise into low-opacity regions by downscaling $\mathbf{O}$ into $\mathbf{O}_z$ through max pooling to match $\mathbf{z}_{deg}$'s spatial dimensions (we retain fine-grained information via \cref{eq:fine-opacity}) and deriving $\mathbf{z}_{mix} = \mathbf{O}_z \mathbf{z}_{deg} + (1 - \mathbf{O}_z)\boldsymbol{\epsilon}$ as the source distribution for our model. As no source information is lost from the max-pooling, this approach preserves the consistency benefits of starting from $\mathbf{z}_{deg}$ while gracefully interpolating to the standard Gaussian prior in entirely novel regions. This strategy is conceptually linked to inpainting methods~\cite{avrahami2022blended, kim2025rad, mayet2025tdpaint} that preserve known regions at low noise while pushing unknown regions toward the generative prior, though we operate with a continuous opacity signal rather than a binary mask. We formally derive compatibility with flow matching in \cref{sec:opacity-mixing-derivation}.

\paragraph{Data curation.} Our goal is to not only correct artifacts in under-observed areas as in prior work~\cite{wu2025difix3d+, fischer2025flowr} but also generate plausible content in entirely unseen areas. To do so, we generate paired reconstruction-ground truth samples from DL3DV-10K~\cite{ling2024dl3dv} with a camera selection strategy that encourages highly sparse reconstructions with large empty regions that the model must learn to inpaint. Given a set of camera poses with rotations $\mathbf{R}_i$ and translations $\mathbf{t}_i$, we first measure the camera pose distance $d_{ij} = \theta_{ij}/\pi + \lambda_t\,\lVert\mathbf{t}_i - \mathbf{t}_j\rVert_2 / \bar r$, where $\theta_{ij}\in[0,\pi]$ is the SO(3) geodesic angle (in radians) between $\mathbf{R}_i$ and $\mathbf{R}_j$, $\bar r = \tfrac{1}{N}\sum_k \lVert\mathbf{t}_k\rVert_2$ is the mean L2 norm of the camera positions in the scene, and $\lambda_t = 1$; this puts both terms on the same order of magnitude (the rotation term lies in $[0,1]$, and translations are normalized to unit mean radius). We then find the camera pair $(P_1, P_2)$ with the largest distance, and seed groups $G_1$ and $G_2$. We assign the remaining cameras to $G_1$ or $G_2$ based on their distance to $P_1$ and $P_2$, and then sample 2-12 cameras with the largest inter-camera distance within each group to generate reconstructions of differing sparsity. We roughly align the camera scales of each reconstruction with a pretrained metric depth estimator~\cite{wang2025moge2} and prompt a vision-language model~\cite{bai2025qwen3vltechnicalreport} for scene descriptions. We provide more details in \cref{sec:sparse-reconstruction} of the supplement.

\paragraph{Optimization.} Given an initial latent-encoded rendering $\mathbf{z}_{deg}$, which we transform into $\mathbf{z}_{mix}$, we train our model to predict its enhanced counterpart $\mathbf{z}_{enh}$ via conditional flow matching loss $\mathcal{L}_{cfm}$~\cite{lipman2023flow}. We construct batches of paired reconstruction-ground truth data by sampling $N=81$ frames along with the corresponding camera poses, text prompt (dropped with 10\% probability), and a uniformly varying number of reference views (0-12). To enhance the model's generative abilities and viewpoint controllability, we drop the last $K \leq N$ frames of the input ($K$ is randomly chosen) by zeroing both the RGB rendering and opacity map while retaining the Plücker raymaps, so that the model must rebuild the ground truth from the prompt, reference views, and camera conditions alone.

\subsection{Causal Distillation}
\label{sec:causal-distillation}

\paragraph{Initialization.} We initialize the causal model from the weights of the bidirectional teacher. To stabilize training, we follow a simpler strategy than the ODE initialization protocol of prior work~\cite{yin2025causvid, huang2025selfforcing, shin2025motionstream}, which requires generating a dataset of ODE trajectories from the teacher model. Instead, we simply apply a block-causal mask, perturb each input frame with differing noise levels as in Diffusion Forcing~\cite{chen2025diffusion}, and otherwise use the same inputs and training protocol as in \cref{sec:bidirectional-training}.

\paragraph{Autoregressive rollout.} After initialization, we adopt a training strategy similar to Self Forcing~\cite{huang2025selfforcing}, where we generate video chunks sequentially and condition on previously generated chunks via KV caching, except that we continue applying dropout as in \cref{sec:bidirectional-training} as camera control and generation from pure noise otherwise degrade. We apply Distribution Matching Distillation (DMD)~\cite{yin2024onestep} to convert the model into a few-step generator ($N=4$ in our experiments, although, outside of entirely novel regions, this can often be reduced to fewer steps with little noticeable difference as discussed in \cref{sec:denoising-steps} of the supplement).

\paragraph{Long video generation.} Existing methods rely on long-horizon training~\cite{yang2025longlive, hong2025relicinteractivevideoworld} to minimize error accumulation in long video rollouts. Although these strategies can be applied to our method, in practice we find our conditioning signals (notably the degraded rendering and reference views) sufficient to prevent error accumulation. We thus train with the same number of frames as in \cref{sec:bidirectional-training} and use a rolling KV cache during inference. 

Although simple, this approach accelerates training convergence (due to training on a more diverse set of shorter videos for a given computational budget) and generalizes to arbitrary length videos, as shown in our experiments.

\begin{table*}[t]
  \centering
  \caption{\textbf{Artifact removal on Nerfbusters and DL3DV}. All \method\ variants outperform prior methods by a considerable margin, improving PSNR by 2 dB.}
  \label{tab:difix_protocol}
  \resizebox{0.9\linewidth}{!}{
  \begin{tabular}{@{}l|cccc|cccc@{}}
    \toprule
    & \multicolumn{4}{c|}{Nerfbusters~\cite{warburg2023nerfbusters}} & \multicolumn{4}{c}{DL3DV~\cite{ling2024dl3dv}} \\
    Method & PSNR$\uparrow$ & SSIM$\uparrow$ & LPIPS$\downarrow$ & FID$\downarrow$ & PSNR$\uparrow$ & SSIM$\uparrow$ & LPIPS$\downarrow$ & FID$\downarrow$ \\
    \midrule
    Nerfacto~\cite{tancik2023nerfstudio} & 17.29 & 0.621 & 0.402 & 134.65 & 17.16 & 0.581 & 0.430 & 112.30 \\
    3DGS~\cite{kerbl3Dgaussians} & 17.66 & 0.678 & 0.327 & 113.84 & 17.18 & 0.588 & 0.384 & 107.23 \\
    Nerfbusters~\cite{warburg2023nerfbusters} & 17.72 & 0.647 & 0.352 & 116.83 & 17.45 & 0.606 & 0.370 & 96.61 \\
    GANeRF~\cite{roessle2023ganerf} & 17.42 & 0.611 & 0.354 & 115.60 & 17.54 & 0.610 & 0.342 & 81.44 \\
    NeRFLiX~\cite{zhou2023nerflix} & 17.91 & 0.656 & 0.346 & 113.59 & 17.56 & 0.610 & 0.359 & 80.65 \\
    \textsc{Difix3D} (Nerfacto)~\cite{wu2025difix3d+} & 18.08 & 0.653 & 0.328 & 63.77 & 17.80 & 0.596 & 0.327 & 50.79 \\
    \textsc{Difix3D} (3DGS)~\cite{wu2025difix3d+} & 18.14 & 0.682 & 0.287 & 51.34 & 17.80 & 0.598 & 0.314 & 50.45 \\    
    \textsc{Difix3D+} (Nerfacto)~\cite{wu2025difix3d+}  & 18.32 & 0.662 & 0.279 & 49.44 & 17.82 & 0.613 & 0.283 & 41.77 \\
    \textsc{Difix3D+} (3DGS)~\cite{wu2025difix3d+} & 18.51 & 0.686 & \cellcolor{orange}0.264 & 41.77 & 17.99 & 0.602 & 0.293 & 40.86 \\
    \hline
    \textbf{\method} & \cellcolor{yellow}19.83 & \cellcolor{yellow}0.701 & \cellcolor{red} 0.254 & \cellcolor{red} 37.78 & \cellcolor{yellow}19.73 & \cellcolor{yellow}0.672 & \cellcolor{red}0.231 & \cellcolor{red}20.85 \\
    \textbf{\methodTD} & \cellcolor{red}20.24 & \cellcolor{red}0.729 & 0.267 & \cellcolor{orange}39.67 & \cellcolor{red}20.14 & \cellcolor{red}0.705 & \cellcolor{yellow}0.256 & \cellcolor{yellow}24.27 \\
    \textbf{\methodTDp} & \cellcolor{orange}20.12 & \cellcolor{orange}0.713 & \cellcolor{orange}0.264 & \cellcolor{yellow} 41.17 & \cellcolor{orange}20.06 & \cellcolor{orange}0.686 & \cellcolor{orange}0.242 & \cellcolor{orange}22.61 \\

    \bottomrule
  \end{tabular}
   }
\end{table*}

\paragraph{3D distillation.} Prior work distills diffusion model outputs into 3D representations~\cite{kerbl3Dgaussians} for consistency purposes, as they otherwise exhibit temporal instability~\cite{wu2025difix3d+} or are limited by number of frames bidirectional models can generate in a single pass~\cite{Wu2025GenFusion, fischer2025flowr}. As our auto-regressive model can sequentially generate arbitrary-length renderings, we are not limited by these constraints. However, 3D distillation is still sometimes desirable from an efficiency perspective, as these representations render orders of magnitude faster. To do so, existing methods require a progressive distillation process that alternates between view generation and 3D reconstruction, incurring significant training time overhead. In our case, as we can generate an arbitrary number of frames in a consistent manner, we adopt a more efficient approach by simply generating all desired novel views in a single pass before applying standard 3D reconstruction.

\section{Experiments}
\label{sec:experiments}
We evaluate three variants of our method: \method, which directly renders novel views from the auto-regressive generator, \methodTD, which distills its outputs back into the underlying 3D representation, and \methodTDp, which re-applies the auto-regressive model as a post-processing step on top of \methodTD (as in \cite{wu2025difix3d+}). We assess their ability to enhance in-the-wild captures against a wide range of prior work in \cref{sec:in-the-wild} and their capacity to synthesize unobserved regions on a more challenging dataset split against a smaller set of relevant baselines in \cref{sec:novel-content}. We validate the contribution of individual components in \cref{sec:diagnostics}.

\begin{table*}[]
    \centering
    \caption{\textbf{Sparse view reconstruction methods on the Mip-NeRF~360 dataset.} We exceed existing work by a wide margin across every metric.}
    \label{tab:genfusion_protocol}
    \resizebox{\linewidth}{!}{
    \begin{tabular}{@{}l@{\,\,}|ccc|ccc|ccc}
    \toprule
    & \multicolumn{3}{c|}{PSNR $\uparrow$} & \multicolumn{3}{c|}{SSIM $\uparrow$} & \multicolumn{3}{c}{LPIPS $\downarrow$}  \\
    Method & 3-view & 6-view & 9-view  & 3-view & 6-view & 9-view  & 3-view & 6-view & 9-view   \\
    \midrule
    Zip-NeRF~\cite{Jonathan2023ICCV} & 12.77 & 13.61 & 14.30 & 0.271 & 0.284 & 0.312 & 0.705 & 0.663 & 0.633  \\
    3DGS~\cite{kerbl3Dgaussians} & 13.06 & 14.96 & 16.79  & 0.251 & 0.355 &  0.447  &  0.576 &  0.505 &  0.446  \\
    2DGS~\cite{Huang2DGS2024} & 13.07 & 15.02 & 16.67 & 0.243 & 0.338 & 0.423  &  0.580 &  0.506 &  0.449 \\
    FSGS~\cite{zhu2023FSGS} & 14.17 &  16.12 &  17.94 &  0.318 &  0.415 &  0.492  &  0.578 & 0.517 & 0.468  \\
    FreeNeRF~\cite{Yang2023FreeNeRF} & 12.87 & 13.35 & 14.59 & 0.260 & 0.283 & 0.319 & 0.715 & 0.717 & 0.695  \\
    SimpleNeRF~\cite{somraj2023simplenerf} & 13.27 & 13.67 & 15.15 & 0.283 & 0.312 & 0.354  & 0.741 & 0.721 & 0.676 \\
    DiffusioNeRF~\cite{DiffusioNeRF} & 11.05 & 12.55 & 13.37  & 0.189 & 0.255 & 0.267  & 0.735 & 0.692 & 0.680  \\
    ZeroNVS~\cite{Kyle2024CVPR} &  14.44 & 15.51 & 15.99  & 0.316 & 0.337 & 0.350 & 0.680 & 0.663 & 0.655 \\
    DNGaussian~\cite{li2024dngaussian} & 14.00 & 15.21 & 16.72 & 0.301 & 0.356 & 0.397 & 0.620 & 0.604 & 0.603 \\
    FlowR~\cite{fischer2025flowr} & 14.46 & 16.18 & 17.53 & 0.347 & 0.409 & 0.456 & 0.587 & 0.520 & 0.467 \\
    ReconFusion~\cite{Rundi2024CVPR} &  15.50 &  16.93 &  18.19 &  0.358 &  0.401 & 0.432 & 0.585 & 0.544 & 0.511  \\
    GenFusion~\cite{Wu2025GenFusion} &  15.29 &  17.16 &  18.36  &  0.369 & 0.447 & 0.496  & 0.585 &  0.500 &  0.465  \\
    GSFixer~\cite{yin2025gsfixerimproving3dgaussian} &  15.61 & 17.27 & 18.63 & 0.370 & 0.426 & 0.481 & 0.559 & 0.478 & 0.420  \\
    CAT3D~\cite{gao2024cat3d} &  16.62 & 17.72 &  18.67 &  0.377 &  0.425 & 0.460  & 0.515 &  0.482 &  0.460  \\
    \hline
    \textbf{\method} & \cellcolor{yellow}17.06 & \cellcolor{orange}18.64 &  \cellcolor{yellow}19.96 & \cellcolor{yellow}0.420  & \cellcolor{yellow}0.476 &  \cellcolor{yellow}0.518 & \cellcolor{red}0.437 &  \cellcolor{orange}0.390 & \cellcolor{orange}0.353  \\
    \textbf{\methodTD} & \cellcolor{orange}17.29 & \cellcolor{red}18.95 &  \cellcolor{red}20.24 & \cellcolor{red}0.451  & \cellcolor{red}0.526 &  \cellcolor{red}0.598 & \cellcolor{orange}0.440 &  \cellcolor{red}0.382 & \cellcolor{red}0.327  \\
    \textbf{\methodTDp} & \cellcolor{red}17.51 & \cellcolor{red}18.95 &  \cellcolor{orange}20.16 & \cellcolor{orange}0.444  & \cellcolor{orange}0.498 &  \cellcolor{orange}0.537 & \cellcolor{yellow}0.441 &  \cellcolor{yellow}0.396 & \cellcolor{yellow}0.359  \\
    \bottomrule
    \end{tabular}
    }
\end{table*}

\subsection{Implementation}

We implement our method in PyTorch~\cite{paszke2019pytorch} and train it on 128 H100 GPUs, using a batch size of one per GPU (128 total). We use FlashAttention-3~\cite{shah2024flashattention3fastaccurateattention} for acceleration. In our main experiments, we finetune the bidirectional model described in \cref{sec:bidirectional-training} for 15,000 iterations using AdamW~\cite{loshchilov2018decoupled} with a learning rate of $1 \times 10^{-5}$. We then initialize the causal model for 5,000 iterations with the same learning rate, followed by 2,000 iterations of auto-regressive rollout and DMD training ($\approx$15k GPU-hours total), using learning rates of $2\times10^{-6}$ for the generator and $4\times10^{-7}$ for the fake score function. For the ablations, we use a truncated schedule of 10,000 + 2,000 + 600 iterations on 64 GPUs to reduce computational cost ($\approx$4k GPU-hours). We use 3DGUT~\cite{wu20253dgut} with MCMC densification~\cite{kheradmand20243d} for the initial reconstructions used by our model. At test time, we use $K\!=\!6$ uniformly sampled reference views for experiments matching the Difix3D+ protocol (\cref{tab:difix_protocol}) and all available input views otherwise (\cref{tab:genfusion_protocol,tab:our_protocol}). We use prompts generated by a vision-language model (\cref{sec:sparse-reconstruction}). Baselines are evaluated following their standard protocols.

\subsection{Enhancing In-the-Wild Captures}
\label{sec:in-the-wild}

\paragraph{Datasets.} We run comparisons on Nerfbusters~\cite{warburg2023nerfbusters} and DL3DV~\cite{ling2024dl3dv} using the splits provided by \cite{wu2025difix3d+}, and on Mip-NeRF~360~\cite{barron2022mipnerf360} with the splits proposed by \cite{Rundi2024CVPR} and used in subsequent work~\cite{gao2024cat3d, Wu2025GenFusion}.

\paragraph{Baselines.} We compare \method\ to an extensive set of baselines, including the original 3DGS~\cite{kerbl3Dgaussians} and 2DGS~\cite{Huang2DGS2024}, NeRF variants~\cite{tancik2023nerfstudio, Jonathan2023ICCV}, non-generative sparse reconstruction methods~\cite{zhu2023FSGS, Yang2023FreeNeRF, somraj2023simplenerf, li2024dngaussian}, and other diffusion-based work~\cite{warburg2023nerfbusters, wu2025difix3d+, DiffusioNeRF, Kyle2024CVPR, Rundi2024CVPR, Wu2025GenFusion, gao2024cat3d, yin2025gsfixerimproving3dgaussian, fischer2025flowr}.

\paragraph{Metrics.} We calculate PSNR, SSIM~\cite{ssim}, LPIPS~\cite{zhang2018perceptual}, and FID~\cite{heusel2017gans} on Nerfbusters and DL3DV using the same protocol and metric implementations as Difix3D+~\cite{wu2025difix3d+}. On Mip-NeRF~360, we calculate PSNR, SSIM, and LPIPS across the 3-, 6-, and 9-view splits using the same implementations as GenFusion~\cite{Wu2025GenFusion}.

\begin{figure}
  \centering
    \includegraphics[width=\linewidth]{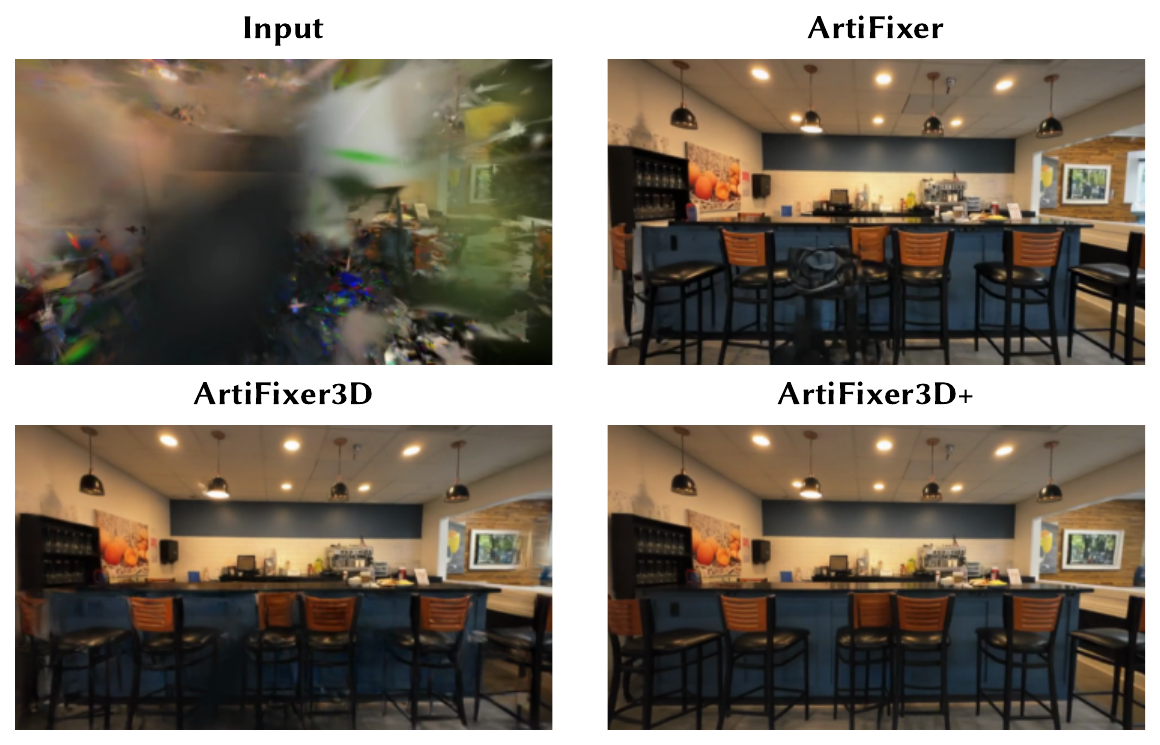}
    \caption{\textbf{\method variants.} Most visible differences occur in highly corrupted regions. \methodTD's explicit 3D consistency improves fidelity with the source images and mitigates transient corruption (\textbf{middle}), at the cost of some sharpness, which \methodTDp restores. Nonetheless, all variants outperform prior work by a substantial margin.}
    \label{fig:artifixer-variants}
\end{figure}

\begin{figure*}
  \centering
    \includegraphics[width=0.8\textwidth]{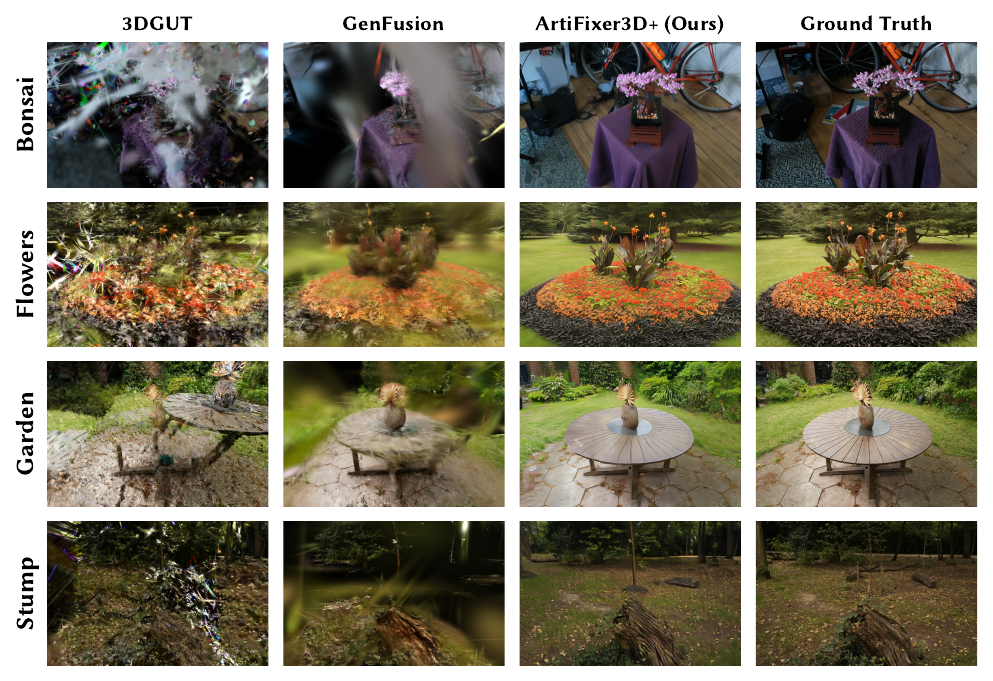}
\caption{\textbf{Mip-NeRF 360 results}. We present visualizations of Mip-NeRF's most challenging split (3-view). Our results far exceed all prior work both quantitatively and qualitatively. Our method is able to recover the correct geometry from the reference views even in scenarios where the input rendering is completely inaccurate (table in \textbf{third row}).}
    \label{fig:m360-qualitative}
    \vspace{-1mm}
\end{figure*}

\begin{figure*}
  \centering
    \includegraphics[width=0.9\textwidth]{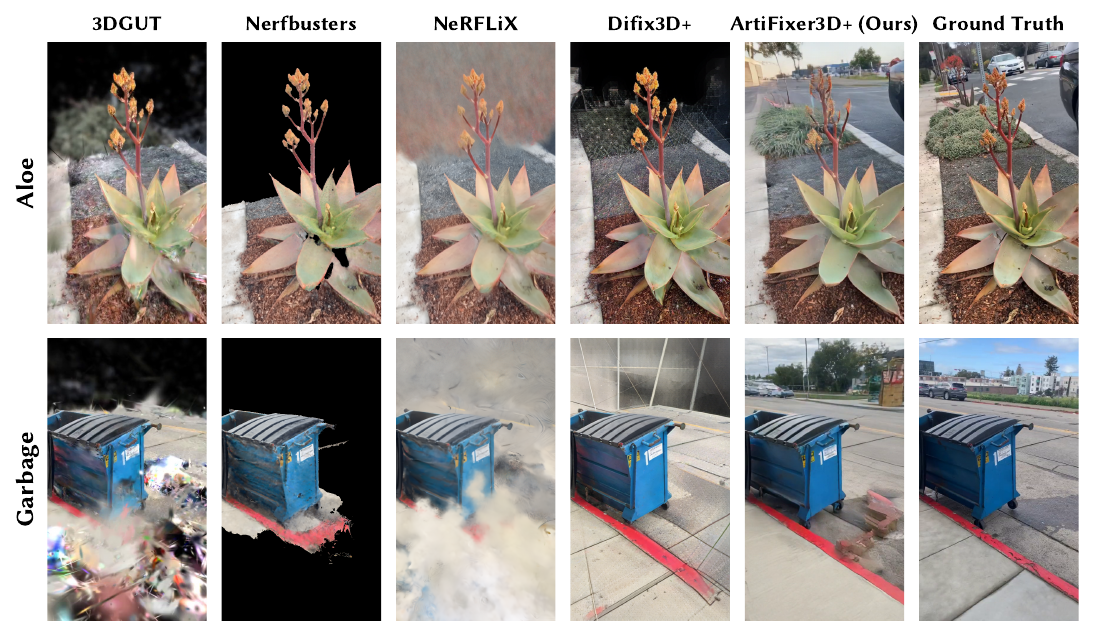}
\caption{\textbf{Nerfbusters results}. As with the other datasets, our method is the only one to generate plausible visuals in unseen regions while preserving the fidelity of the original content.}
    \label{fig:nerfbusters}
    \vspace{-4mm}
\end{figure*}

\paragraph{Results.} We present quantitative results for Nerfbusters and DL3DV in \cref{tab:difix_protocol} and Mip-NeRF 360 in \cref{tab:genfusion_protocol}. We provide visual comparisons in \cref{fig:m360-qualitative} and \cref{fig:nerfbusters}. All \method\ variants outperform all baselines by a substantial margin. Although the different variants produce similar renderings, \method's are slightly sharper, while \methodTD's are even more consistent with the source images at the cost of some blurriness due to its explicit 3D representation, leading to a minor increase in PSNR and SSIM and a small degradation in LPIPS and FID in \cref{tab:difix_protocol}. Re-applying the generator to the improved 3D reconstruction (\methodTDp) restores some of this sharpness, leading to renderings that are crisper than \methodTD and slightly more consistent than \method (\cref{fig:artifixer-variants}).

\begin{table}[t]
  \centering
  \caption{\textbf{Novel content generation.} We reconstruct DL3DV scenes following a protocol that creates large areas unobserved by training views. We outperform the next-best method (GenFusion) by almost 3 dB in PSNR.}
  \label{tab:our_protocol}
  \resizebox{\linewidth}{!}{
  \begin{tabular}{@{}l|cccccc@{}}
    \toprule
    Method & PSNR$\uparrow$ & SSIM$\uparrow$ & LPIPS$\downarrow$ & FID$\downarrow$ \\
    \midrule
    3DGUT~\cite{wu20253dgut} & 16.12 & 0.537 & 0.445 & 92.94 \\
    \textsc{Difix3D} (Nerfacto)~\cite{wu2025difix3d+} & 14.16 & 0.453 & 0.545 & 74.59 \\
    \textsc{Difix3D} (3DGS)~\cite{wu2025difix3d+} & 16.60 & 0.599 & 0.405 & 52.70 & \\
    \textsc{Difix3D+} (Nerfacto)~\cite{wu2025difix3d+} & 13.74 & 0.434 & 0.483 & 30.07 \\
    \textsc{Difix3D+} (3DGS)~\cite{wu2025difix3d+} & 16.34 & 0.564 & 0.382 & 21.77 \\
    Fixer (offline)~\cite{nvidia-fixer} & 13.09 & 0.355 & 0.584 & 135.43  \\
    Fixer (online)~\cite{nvidia-fixer} & 13.93 & 0.443 & 0.535 & 79.44 \\
    Gen3C~\cite{ren2025gen3c} & 15.50 & 0.491 & 0.476 & 68.36 & \\
    GenFusion~\cite{Wu2025GenFusion} & 17.03 & 0.624 & 0.392 & 132.91 & \\
    \hline
    \textbf{\method} & \cellcolor{yellow}19.75 & \cellcolor{yellow}0.643 & \cellcolor{red}0.303 & \cellcolor{red}12.22 \\
    \textbf{\methodTD} & \cellcolor{orange}19.92 & \cellcolor{red}0.673 & \cellcolor{orange}0.306 & \cellcolor{yellow}16.28 \\
    \textbf{\methodTDp} & \cellcolor{red}20.15 & \cellcolor{orange}0.662 & \cellcolor{yellow}0.307 & \cellcolor{orange}13.91 \\
    \bottomrule
  \end{tabular}
  }
\end{table}

\subsection{Novel Content Generation}
\label{sec:novel-content}

\paragraph{Dataset.} We evaluate novel content generation by following the sparse reconstruction protocol described in \cref{sec:sparse-reconstruction} on scenes from DL3DV, resulting in numerous ``holes" that must be corrected in a manner consistent with existing observations.

\paragraph{Baselines.} We compare to a smaller set of baselines most relevant to our work, notably 3DGUT~\cite{wu20253dgut} as the base representation we provide as initial renderings to our method, image-based diffusion methods via Difix3D+~\cite{wu2025difix3d+} and Fixer~\cite{nvidia-fixer}, and approaches that build upon bidirectional video models~\cite{Wu2025GenFusion, ren2025gen3c}.

\begin{figure*}
  \centering
    \includegraphics[width=\textwidth]{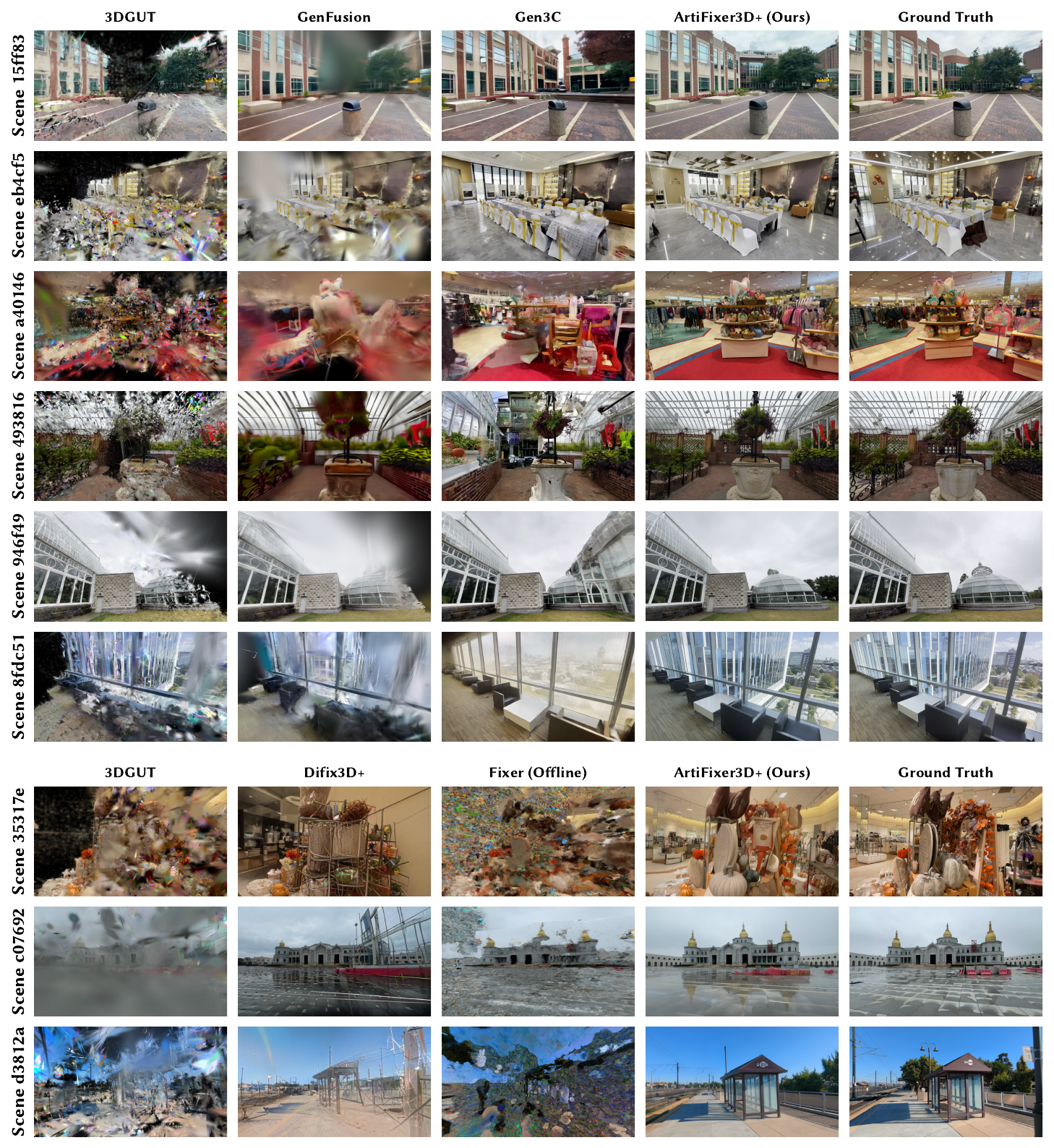}
    \vspace{-4mm}
\caption{\textbf{DL3DV results}. We compare \methodTDp to its initial 3DGUT~\cite{wu20253dgut} input, two baselines that build upon bidirectional video diffusion models (\textbf{top rows}), and two that leverage image models (\textbf{bottom rows}). GenFusion~\cite{Wu2025GenFusion}'s video model generates 16 frames at a time, requiring an iterative distillation process that leads to blurry results, especially in empty areas. Gen3C~\cite{ren2025gen3c}'s renderings are sharper but often do not respect the source content (background in \textbf{top row}), have incorrect geometry (\textbf{second row}), and exhibit color shift (\textbf{sixth row}). Methods that directly take renderings as input without opacity mixing~\cite{wu2025difix3d+, nvidia-fixer} fail to reconstruct empty regions. Our method can reconstruct plausible and consistent geometry even when the initial rendering is highly degraded. Please refer to our project website for comparison videos.}
    \label{fig:our-protocol-qualitative}
    \vspace{-4mm}
\end{figure*}

\paragraph{Results.} We present quantitative results, using the same metrics as \cref{tab:difix_protocol}, in \cref{tab:our_protocol}. We provide qualitative results in \cref{fig:our-protocol-qualitative}. All \method\ variants outperform the next-best method (GenFusion~\cite{Wu2025GenFusion}) by almost 3 dB in PSNR. Gen3C~\cite{ren2025gen3c} gives the next-best visually appealing results, but its conditioning often does not respect the source content, and its quality is upper-bounded by the depth estimator it uses to generate its 3D cache (in contrast to our purely data-driven approach). Difix3D+~\cite{wu2025difix3d+} and Fixer~\cite{nvidia-fixer} generally fail to inpaint plausible context due to their deterministic conditioning.

\begin{table}[t]
  \centering
  \caption{\textbf{Diagnostics.} We evaluate reconstruction quality on Mip-NeRF 360. Denoising input renderings instead of conditioning via channel concatenation is crucial to producing outputs consistent with source images.}
  \label{tab:diagnostics}
\resizebox{\linewidth}{!}{
\begin{tabular}{l|ccc|cccc}
\toprule
Method & 
\makecell{Direct \\ Input} &
\makecell{Opacity \\ Mixing} &
\makecell{Diffusion \\ Forcing} &
PSNR$\uparrow$ &
SSIM$\uparrow$ &
LPIPS$\downarrow$ &
FID$\downarrow$ \\ \midrule

Channel Concatenation & \xmark & \xmark & \textcolor{CheckGreen}{\checkmark}  & 14.52 & 0.391 & 0.490 & 87.551 \\
w/o Opacity Mixing & \textcolor{CheckGreen}{\checkmark}& \xmark & \textcolor{CheckGreen}{\checkmark} &  \cellcolor{yellow}17.34 & \cellcolor{yellow}0.440 & \cellcolor{yellow}0.429 & \cellcolor{yellow}87.058 \\
w/o Initialization & \textcolor{CheckGreen}{\checkmark} & \textcolor{CheckGreen}{\checkmark} & \xmark & \cellcolor{orange}17.58 & \cellcolor{orange}0.450 & \cellcolor{orange}0.416 & \cellcolor{orange}74.924 \\
\midrule
Full Method & \textcolor{CheckGreen}{\checkmark} & \textcolor{CheckGreen}{\checkmark} & \textcolor{CheckGreen}{\checkmark} & \cellcolor{red}17.99 & \cellcolor{red}0.461 & \cellcolor{red}0.408 & \cellcolor{red}69.43 \\
\end{tabular}
}
\end{table}

\subsection{Diagnostics}
\label{sec:diagnostics}

\paragraph{Ablations.} We ablate the effectiveness of our opacity mixing strategy by comparing it to variants that instead use channel concatenation or omit the opacity mixing. We also measure the impact of the causal model weight initialization described in \cref{sec:causal-distillation}. We report results on the Mip-NeRF 360 dataset averaged over all splits in \cref{tab:diagnostics} and show that our design choice of starting from the initial rendering instead of conditioning on it via channel concatenation is essential to rendering consistently with the source imagery. Our causal initialization method is not essential as the model still converges to a competitive level of quality, but provides a modest boost.

\begin{figure}
  \centering
    \includegraphics[width=\linewidth]{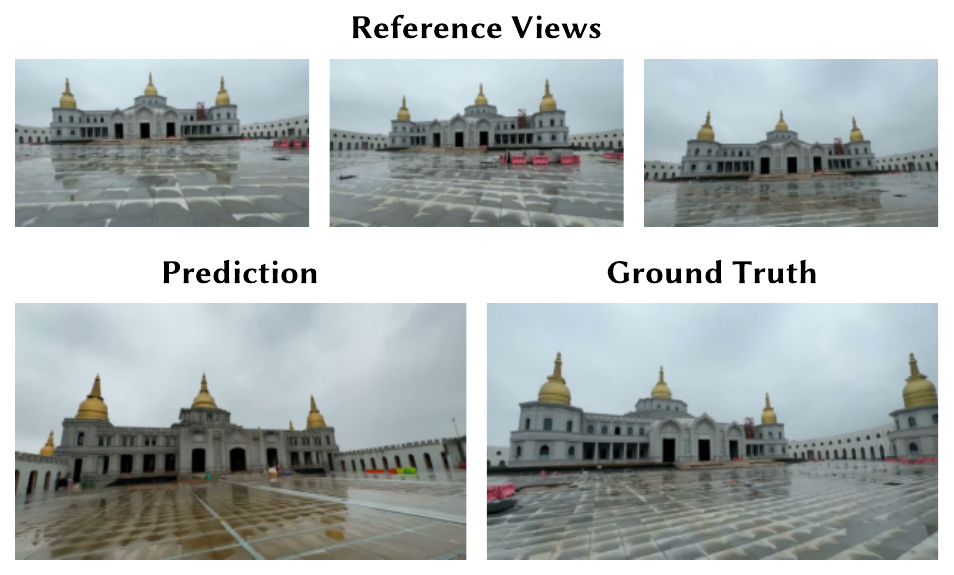}
    \caption{\textbf{Reference views.} Without the initial rendering condition, \method can generate predictions from the reference views. Although fidelity drops somewhat, the high-level structure of the scene remains intact.}
    \label{fig:reference-views}
\end{figure}

\begin{figure}
  \centering
    \includegraphics[width=\linewidth]{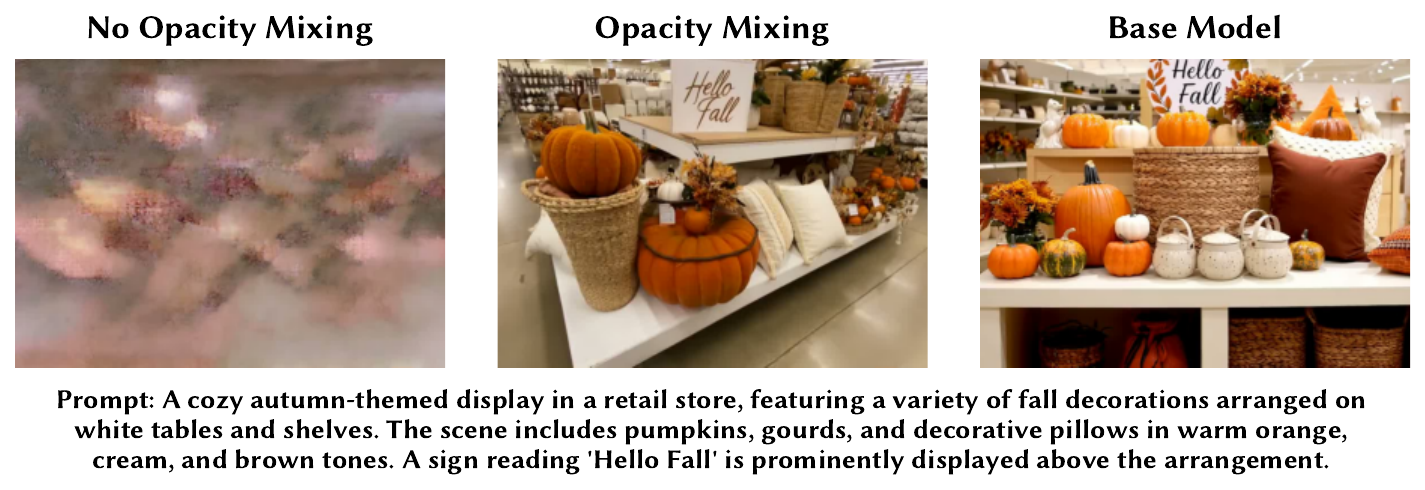}
    \caption{\textbf{Text-to-video generation.} To illustrate our model's generative ability, we generate videos from text prompts alone. With opacity mixing, it retains similar quality to its base model~\cite{wan2025}.}
    \label{fig:t2v}
\end{figure}

\paragraph{Conditioning.} To probe which inputs drive output quality, we progressively strip conditioning signals. First, we drop the initial rendering, forcing the model to rely solely on reference views and camera rays. Although fidelity decreases, the model still recovers the high-level scene structure (\cref{fig:reference-views}). Next, we remove all conditioning except the text prompt, reverting to standard text-to-video generation; output quality remains comparable to the base Wan 2.1 model (\cref{fig:t2v}).

\begin{table}
    \centering
    \caption{\textbf{Inference speed.} Causal distillation yields a $70\times$ speedup over the bidirectional Wan 2.1 backbones. \methodTD renders directly from 3DGUT. Additional configurations are reported in \cref{tab:timing-full}.}
    \label{tab:timing}
    \resizebox{0.75\linewidth}{!}{
    \begin{tabular}{@{}l|c@{}}
      \toprule
      Method & FPS $\uparrow$ \\
      \midrule
      Wan 2.1 T2V-14B~\cite{wan2025} & 0.12 \\
      Wan 2.1 T2V-1.3B~\cite{wan2025} & 0.49 \\
      \hline
      \textbf{\method\!/\!\methodTDp (14B)} & \cellcolor{yellow}8.36 \\
      \textbf{\method\!/\!\methodTDp (1.3B)} & \cellcolor{orange}34.38 \\
      \textbf{\methodTD} & \cellcolor{red}268 \\
      \bottomrule
    \end{tabular}
    }
\end{table}

\paragraph{Model scale.} To disentangle model scale from our other contributions, we train with Wan 2.1 T2V-1.3B and report results in \cref{sec:additional-experiments}.

\paragraph{Timing.} We report inference speed in \cref{tab:timing} on a single GB300 GPU. Causal distillation with KV caching and few-step sampling yields a $70\times$ speedup over the bidirectional Wan 2.1 14B and 1.3B backbones. With the 14B backbone, \method and \methodTDp reach 8.36 FPS. Our 1.3B variant reaches 34.38 FPS. \methodTD renders at native 3DGUT speed (268 FPS). Fewer denoising steps and context parallelism provide further gains (\cref{sec:denoising-steps}).

\section{Conclusion}
\label{sec:conclusion}
Neural reconstruction and camera-controlled video generation provide complementary strengths for novel view synthesis. In this work, we introduced \method, an auto-regressive video diffusion model that seeks to combine the advantages of both paradigms. \method transforms corrupted renderings of reconstructed scenes into clean, temporally consistent frames, while retaining sufficient generative capacity to inpaint unobserved regions and the efficiency required for interactive use. The strong conditioning signal from the reconstructed scene significantly simplifies distillation and conversion to an auto-regressive formulation, enabling \method to generate long video sequences with less quality degradation.

\section{Acknowledgments}

We thank Zian Wang and Nicholas Sharp for their helpful advice and feedback throughout this project.

\bibliographystyle{ACM-Reference-Format}
\bibliography{main}

@String(CVPR= {IEEE Conf. Comput. Vis. Pattern Recog.})

@String(ICCV= {Int. Conf. Comput. Vis.})

@String(ECCV= {Eur. Conf. Comput. Vis.})

@String(TOG= {ACM Trans. Graph.})

@String(ICLR = {Int. Conf. Learn. Represent.})

@String(CVPR  = {CVPR})

@String(ICCV  = {ICCV})

@String(ECCV  = {ECCV})

@String(TOG   = {ACM TOG})

@String(ICLR  = {ICLR})

@inproceedings{wu2025difix3d+,
  title={DIFIX3D+: Improving 3D Reconstructions with Single-Step Diffusion Models},
  author={Wu, Jay Zhangjie and Zhang, Yuxuan and Turki, Haithem and Ren, Xuanchi and Gao, Jun and Shou, Mike Zheng and Fidler, Sanja and Gojcic, Zan and Ling, Huan},
  booktitle={CVPR},
  pages={26024--26035},
  year={2025}
}

@InProceedings{fischer2025flowr,
    author    = {Tobias Fischer and Samuel Rota Bul{\`o} and Yung-Hsu Yang and Nikhil Keetha and Lorenzo Porzi and Norman M\"uller and Katja Schwarz and Jonathon Luiten and Marc Pollefeys and Peter Kontschieder},
    title     = {{FlowR}: Flowing from Sparse to Dense 3D Reconstructions},
    booktitle = {ICCV},
    year      = {2025}
}

@Article{kerbl3Dgaussians,
      author       = {Kerbl, Bernhard and Kopanas, Georgios and Leimk{\"u}hler, Thomas and Drettakis, George},
      title        = {3D Gaussian Splatting for Real-Time Radiance Field Rendering},
      journal      = {ACM Transactions on Graphics},
      number       = {4},
      volume       = {42},
      month        = {July},
      year         = {2023},
      url          = {https://repo-sam.inria.fr/fungraph/3d-gaussian-splatting/}
}

@inproceedings{ren2025gen3c,
    title={GEN3C: 3D-Informed World-Consistent Video Generation with Precise Camera Control},
    author={Ren, Xuanchi and Shen, Tianchang and Huang, Jiahui and Ling, Huan and 
        Lu, Yifan and Nimier-David, Merlin and M\"uller, Thomas and Keller, Alexander and 
        Fidler, Sanja and Gao, Jun},
    booktitle={CVPR},
    year={2025}
}

@InProceedings{Niemeyer2021Regnerf,
    author    = {Michael Niemeyer and Jonathan T. Barron and Ben Mildenhall and Mehdi S. M. Sajjadi and Andreas Geiger and Noha Radwan},  
    title     = {RegNeRF: Regularizing Neural Radiance Fields for View Synthesis from Sparse Inputs},
    booktitle = {CVPR},
    year      = {2022},
}

@InProceedings{Yang2023FreeNeRF,
    author    = {Jiawei Yang and Marco Pavone and Yue Wang},  
    title     = {FreeNeRF: Improving Few-shot Neural Rendering with Free Frequency Regularization},
    booktitle = {CVPR},
    year      = {2023},
}

@inproceedings{somraj2023simplenerf,
    title = {{SimpleNeRF}: Regularizing Sparse Input Neural Radiance Fields with Simpler Solutions},
    author = {Somraj, Nagabhushan and Karanayil, Adithyan and Soundararajan, Rajiv},
    booktitle = {SIGGRAPH Asia},
    month = {December},
    year = {2023},
    doi = {10.1145/3610548.3618188}
}

@inproceedings{deng2022depth,
  title={Depth-supervised nerf: Fewer views and faster training for free},
  author={Deng, Kangle and Liu, Andrew and Zhu, Jun-Yan and Ramanan, Deva},
  booktitle={CVPR},
  pages={12882--12891},
  year={2022}
}

@inproceedings{roessle2022dense,
  title={Dense depth priors for neural radiance fields from sparse input views},
  author={Roessle, Barbara and Barron, Jonathan T and Mildenhall, Ben and Srinivasan, Pratul P and Nie{\ss}ner, Matthias},
  booktitle={CVPR},
  pages={12892--12901},
  year={2022}
}

@inproceedings{wang2023sparsenerf,
  title={Sparsenerf: Distilling depth ranking for few-shot novel view synthesis},
  author={Wang, Guangcong and Chen, Zhaoxi and Loy, Chen Change and Liu, Ziwei},
  booktitle={ICCV},
  pages={9065--9076},
  year={2023}
}

@InProceedings{zhu2023FSGS, 
title={FSGS: Real-Time Few-Shot View Synthesis using Gaussian Splatting}, 
author={Zehao Zhu and Zhiwen Fan and Yifan Jiang and Zhangyang Wang}, 
booktitle = {ECCV},
year={2024},
}

@inproceedings{yu2022monosdf,
  title={Monosdf: Exploring monocular geometric cues for neural implicit surface reconstruction},
  author={Yu, Zehao and Peng, Songyou and Niemeyer, Michael and Sattler, Torsten and Geiger, Andreas},
  journal={NeurIPS},
  volume={35},
  pages={25018--25032},
  year={2022}
}

@article{song2020score,
  title={Score-based generative modeling through stochastic differential equations},
  author={Song, Yang and Sohl-Dickstein, Jascha and Kingma, Diederik P and Kumar, Abhishek and Ermon, Stefano and Poole, Ben},
  journal={arXiv preprint arXiv:2011.13456},
  year={2020}
}

@inproceedings{zhou2023nerflix,
  title={NeRFLix: High-quality neural view synthesis by learning a degradation-driven inter-viewpoint mixer},
  author={Zhou, Kun and Li, Wenbo and Wang, Yi and Hu, Tao and Jiang, Nianjuan and Han, Xiaoguang and Lu, Jiangbo},
  booktitle={CVPR},
  pages={12363--12374},
  year={2023}
}

@inproceedings{yu2020pixelnerf,
      title={{pixelNeRF}: Neural Radiance Fields from One or Few Images},
      author={Alex Yu and Vickie Ye and Matthew Tancik and Angjoo Kanazawa},
      year={2021},
      booktitle={CVPR},
}

@inproceedings{mvsnerf,
              title={Mvsnerf: Fast generalizable radiance field reconstruction from multi-view stereo},
              author={Chen, Anpei and Xu, Zexiang and Zhao, Fuqiang and Zhang, Xiaoshuai and Xiang, Fanbo and Yu, Jingyi and Su, Hao},
              booktitle={ICCV},
              pages={14124--14133},
              year={2021}
}

@article{kong2025worldwarp,
  title={WorldWarp: Propagating 3D Geometry with Asynchronous Video Diffusion},
  author={Kong, Hanyang and Yang, Xingyi and Zheng, Xiaoxu and Wang, Xinchao},
  journal={arXiv preprint arXiv:2512.19678},
  year={2025}
}

@article{liu2025rolling,
  title={Rolling forcing: Autoregressive long video diffusion in real time},
  author={Liu, Kunhao and Hu, Wenbo and Xu, Jiale and Shan, Ying and Lu, Shijian},
  journal={arXiv preprint arXiv:2509.25161},
  year={2025}
}

@article{cui2025self,
  title={Self-Forcing++: Towards Minute-Scale High-Quality Video Generation},
  author={Cui, Justin and Wu, Jie and Li, Ming and Yang, Tao and Li, Xiaojie and Wang, Rui and Bai, Andrew and Ban, Yuanhao and Hsieh, Cho-Jui},
  journal={arXiv preprint arXiv:2510.02283},
  year={2025}
}

@article{blattmann2023stable,
  title={Stable video diffusion: Scaling latent video diffusion models to large datasets},
  author={Blattmann, Andreas and Dockhorn, Tim and Kulal, Sumith and Mendelevitch, Daniel and Kilian, Maciej and Lorenz, Dominik and Levi, Yam and English, Zion and Voleti, Vikram and Letts, Adam and others},
  journal={arXiv preprint arXiv:2311.15127},
  year={2023}
}

@inproceedings{lu2025infinicube,
  title={InfiniCube: Unbounded and Controllable Dynamic 3D Driving Scene Generation with World-Guided Video Models},
  author={Lu, Yifan and Ren, Xuanchi and Yang, Jiawei and Shen, Tianchang and Wu, Zhangjie and Gao, Jun and Wang, Yue and Chen, Siheng and Chen, Mike and Fidler, Sanja and others},
booktitle={ICCV},
  year={2025}
}

@inproceedings{ren2024scube,
  title={SCube: Instant Large-Scale Scene Reconstruction using VoxSplats},
  author={Ren, Xuanchi and Lu, Yifan and Liang, Hanxue and Wu, Jay Zhangjie and 
    Ling, Huan and Chen, Mike and Fidler, Sanja annd Williams, Francis and Huang, Jiahui},
  booktitle={NeurIPS},
  year={2024},
}

@inproceedings{mildenhall2020nerf,
     title={{NeRF}: Representing Scenes as Neural Radiance Fields for View Synthesis},
     author={Ben Mildenhall and Pratul P. Srinivasan and Matthew Tancik and Jonathan T. Barron and Ravi Ramamoorthi and Ren Ng},
     year={2020},
     booktitle={ECCV},
}

@article{yu2024viewcrafter,
  title={Viewcrafter: Taming video diffusion models for high-fidelity novel view synthesis},
  author={Yu, Wangbo and Xing, Jinbo and Yuan, Li and Hu, Wenbo and Li, Xiaoyu and Huang, Zhipeng and Gao, Xiangjun and Wong, Tien-Tsin and Shan, Ying and Tian, Yonghong},
  journal={arXiv preprint arXiv:2409.02048},
  year={2024}
}

@inproceedings{warburg2023nerfbusters,
  title={Nerfbusters: Removing ghostly artifacts from casually captured nerfs},
  author={Warburg, Frederik and Weber, Ethan and Tancik, Matthew and Holynski, Aleksander and Kanazawa, Angjoo},
  booktitle={Proceedings of the IEEE/CVF International Conference on Computer Vision},
  pages={18120--18130},
  year={2023}
}

@inproceedings{tancik2023nerfstudio,
  title={Nerfstudio: A modular framework for neural radiance field development},
  author={Tancik, Matthew and Weber, Ethan and Ng, Evonne and Li, Ruilong and Yi, Brent and Wang, Terrance and Kristoffersen, Alexander and Austin, Jake and Salahi, Kamyar and Ahuja, Abhik and others},
  booktitle={ACM SIGGRAPH 2023 Conference Proceedings},
  pages={1--12},
  year={2023}
}

@article{roessle2023ganerf,
  title={Ganerf: Leveraging discriminators to optimize neural radiance fields},
  author={Roessle, Barbara and M{\"u}ller, Norman and Porzi, Lorenzo and Bul{\`o}, Samuel Rota and Kontschieder, Peter and Nie{\ss}ner, Matthias},
  journal={ACM Transactions on Graphics (TOG)},
  volume={42},
  number={6},
  pages={1--14},
  year={2023},
  publisher={ACM New York, NY, USA}
}

@inproceedings{ling2024dl3dv,
  title={Dl3dv-10k: A large-scale scene dataset for deep learning-based 3d vision},
  author={Ling, Lu and Sheng, Yichen and Tu, Zhi and Zhao, Wentian and Xin, Cheng and Wan, Kun and Yu, Lantao and Guo, Qianyu and Yu, Zixun and Lu, Yawen and others},
  booktitle={CVPR},
  pages={22160--22169},
  year={2024}
}

@misc{nvidia-fixer,
  title = {NVIDIA Fixer},
  author = {NVIDIA},
  howpublished = {\url{https://huggingface.co/nvidia/Fixer}},
  note = {Accessed: 2026-01-26},
  year={2025}
}

@inproceedings{Wu2025GenFusion,
    author = {Sibo Wu and Congrong Xu and Binbin Huang and Geiger Andreas and Anpei Chen},
    title = {GenFusion: Closing the Loop between Reconstruction and Generation via Videos},
    booktitle = {CVPR},
    year = {2025}
}

@inproceedings{Jonathan2023ICCV,
  author       = {Jonathan T. Barron and
                  Ben Mildenhall and
                  Dor Verbin and
                  Pratul P. Srinivasan and
                  Peter Hedman},
  title        = {Zip-NeRF: Anti-Aliased Grid-Based Neural Radiance Fields},
  booktitle = {ICCV},
  year         = {2023}
}

@inproceedings{DiffusioNeRF,
  author       = {Jamie Wynn and
                  Daniyar Turmukhambetov},
  title        = {DiffusioNeRF: Regularizing Neural Radiance Fields with Denoising Diffusion
                  Models},
  booktitle = {CVPR},
  year         = {2023},
}

@inproceedings{Kyle2024CVPR,
  author       = {Kyle Sargent and
                  Zizhang Li and
                  Tanmay Shah and
                  Charles Herrmann and
                  Hong{-}Xing Yu and
                  Yunzhi Zhang and
                  Eric Ryan Chan and
                  Dmitry Lagun and
                  Li Fei{-}Fei and
                  Deqing Sun and
                  Jiajun Wu},
  title        = {ZeroNVS: Zero-Shot 360-Degree View Synthesis from a Single Image},
  booktitle    = {CVPR},
  year         = {2024},
}

@inproceedings{Rundi2024CVPR,
  author       = {Rundi Wu and
                  Ben Mildenhall and
                  Philipp Henzler and
                  Keunhong Park and
                  Ruiqi Gao and
                  Daniel Watson and
                  Pratul P. Srinivasan and
                  Dor Verbin and
                  Jonathan T. Barron and
                  Ben Poole and
                  Aleksander Holynski},
  title        = {ReconFusion: 3D Reconstruction with Diffusion Priors},
  booktitle    = {CVPR},
  year         = {2024},
}

@inproceedings{Huang2DGS2024,
    title={2D Gaussian Splatting for Geometrically Accurate Radiance Fields},
    author={Huang, Binbin and Yu, Zehao and Chen, Anpei and Geiger, Andreas and Gao, Shenghua},
    booktitle = {SIGGRAPH Asia},
    year      = {2024},
}

@inproceedings{wu20253dgut,
    title={3DGUT: Enabling Distorted Cameras and Secondary Rays in Gaussian Splatting},
    author={Wu, Qi and Martinez Esturo, Janick and Mirzaei, Ashkan and Moenne-Loccoz, Nicolas and Gojcic, Zan},
    booktitle={CVPR},
    year={2025}
}

@inproceedings{poole2023dreamfusion,
title={DreamFusion: Text-to-3D using 2D Diffusion},
author={Ben Poole and Ajay Jain and Jonathan T. Barron and Ben Mildenhall},
booktitle={ICLR},
year={2023},
}

@inproceedings{gao2024cat3d,
    title={CAT3D: Create Anything in 3D with Multi-View Diffusion Models},
    author={Ruiqi Gao* and Aleksander Holynski* and Philipp Henzler and Arthur Brussee and Ricardo Martin-Brualla and Pratul P. Srinivasan and Jonathan T. Barron and Ben Poole*
    },
    journal={NeurIPS},
    year={2024}
}

@inproceedings{liu2023deceptive,
  title={Deceptive-nerf: Enhancing nerf reconstruction using pseudo-observations from diffusion models},
  author={Liu, Xinhang and Chen, Jiaben and Kao, Shiu-hong and Tai, Yu-Wing and Tang, Chi-Keung},
  journal={ECCV},
  year={2024}
}

@inproceedings{liu20243dgs,
  title={3DGS-Enhancer: Enhancing Unbounded 3D Gaussian Splatting with View-consistent 2D Diffusion Priors},
  author={Liu, Xi and Zhou, Chaoyi and Huang, Siyu},
  journal={NeurIPS},
  year={2022}
}

@inproceedings{li2025cameras,
    title={Cameras as Relative Positional Encoding},
    author={Li, Ruilong and Yi, Brent and Liu, Junchen and Gao, Hang and Ma, Yi and Kanazawa, Angjoo},
    journal={NeurIPS},
    year={2025}
}

@article{wan2025,
      title={Wan: Open and Advanced Large-Scale Video Generative Models}, 
      author={Team Wan and Ang Wang and Baole Ai and Bin Wen and Chaojie Mao and Chen-Wei Xie and Di Chen and Feiwu Yu and Haiming Zhao and Jianxiao Yang and Jianyuan Zeng and Jiayu Wang and Jingfeng Zhang and Jingren Zhou and Jinkai Wang and Jixuan Chen and Kai Zhu and Kang Zhao and Keyu Yan and Lianghua Huang and Mengyang Feng and Ningyi Zhang and Pandeng Li and Pingyu Wu and Ruihang Chu and Ruili Feng and Shiwei Zhang and Siyang Sun and Tao Fang and Tianxing Wang and Tianyi Gui and Tingyu Weng and Tong Shen and Wei Lin and Wei Wang and Wei Wang and Wenmeng Zhou and Wente Wang and Wenting Shen and Wenyuan Yu and Xianzhong Shi and Xiaoming Huang and Xin Xu and Yan Kou and Yangyu Lv and Yifei Li and Yijing Liu and Yiming Wang and Yingya Zhang and Yitong Huang and Yong Li and You Wu and Yu Liu and Yulin Pan and Yun Zheng and Yuntao Hong and Yupeng Shi and Yutong Feng and Zeyinzi Jiang and Zhen Han and Zhi-Fan Wu and Ziyu Liu},
      journal = {arXiv preprint arXiv:2503.20314},
      year={2025}
}

@inproceedings{huang2025selfforcing,
  title={Self Forcing: Bridging the Train-Test Gap in Autoregressive Video Diffusion},
  author={Huang, Xun and Li, Zhengqi and He, Guande and Zhou, Mingyuan and Shechtman, Eli},
  journal={NeurIPS},
  year={2025}
}

@article{paszke2019pytorch,
  title={Pytorch: An imperative style, high-performance deep learning library},
  author={Paszke, Adam and Gross, Sam and Massa, Francisco and Lerer, Adam and Bradbury, James and Chanan, Gregory and Killeen, Trevor and Lin, Zeming and Gimelshein, Natalia and Antiga, Luca and others},
  journal={Advances in neural information processing systems},
  volume={32},
  year={2019}
}

@misc{yin2025gsfixerimproving3dgaussian,
      title={GSFixer: Improving 3D Gaussian Splatting with Reference-Guided Video Diffusion Priors}, 
      author={Xingyilang Yin and Qi Zhang and Jiahao Chang and Ying Feng and Qingnan Fan and Xi Yang and Chi-Man Pun and Huaqi Zhang and Xiaodong Cun},
      year={2025},
      eprint={2508.09667},
      archivePrefix={arXiv},
      primaryClass={cs.CV},
      url={https://arxiv.org/abs/2508.09667}, 
}

@inproceedings{lipman2023flowmatching,
    author={Yaron Lipman and Chen, \{Ricky T.Q.\} and Heli Ben-Hamu and Maximilian Nickel and Matt Le},
    title={Flow Matching for Generative Modeling},
    booktitle={ICLR},
    year={2023} 
}

@inproceedings{liu2023flow,
    author={Xingchao Liu and Chengyue Gong and Qiang Liu},
    title={Flow Straight and Fast: Learning to Generate and Transfer Data with Rectified Flow},
    booktitle={ICLR},
    year={2023} 
}

@article{kheradmand20243d,
  title={3d gaussian splatting as markov chain monte carlo},
  author={Kheradmand, Shakiba and Rebain, Daniel and Sharma, Gopal and Sun, Weiwei and Tseng, Yang-Che and Isack, Hossam and Kar, Abhishek and Tagliasacchi, Andrea and Yi, Kwang Moo},
  journal={Advances in Neural Information Processing Systems},
  volume={37},
  pages={80965--80986},
  year={2024}
}

@inproceedings{wang2025moge2,
    title={MoGe-2: Accurate Monocular Geometry with Metric Scale and Sharp Details}, 
    author={Ruicheng Wang and Sicheng Xu and Yue Dong and Yu Deng and Jianfeng Xiang and Zelong Lv and Guangzhong Sun and Xin Tong and Jiaolong Yang},
    booktitle={CVPR},
    year={2025} 
}

@misc{bai2025qwen3vltechnicalreport,
      title={Qwen3-VL Technical Report}, 
      author={Shuai Bai and Yuxuan Cai and Ruizhe Chen and Keqin Chen and Xionghui Chen and Zesen Cheng and Lianghao Deng and Wei Ding and Chang Gao and Chunjiang Ge and Wenbin Ge and Zhifang Guo and Qidong Huang and Jie Huang and Fei Huang and Binyuan Hui and Shutong Jiang and Zhaohai Li and Mingsheng Li and Mei Li and Kaixin Li and Zicheng Lin and Junyang Lin and Xuejing Liu and Jiawei Liu and Chenglong Liu and Yang Liu and Dayiheng Liu and Shixuan Liu and Dunjie Lu and Ruilin Luo and Chenxu Lv and Rui Men and Lingchen Meng and Xuancheng Ren and Xingzhang Ren and Sibo Song and Yuchong Sun and Jun Tang and Jianhong Tu and Jianqiang Wan and Peng Wang and Pengfei Wang and Qiuyue Wang and Yuxuan Wang and Tianbao Xie and Yiheng Xu and Haiyang Xu and Jin Xu and Zhibo Yang and Mingkun Yang and Jianxin Yang and An Yang and Bowen Yu and Fei Zhang and Hang Zhang and Xi Zhang and Bo Zheng and Humen Zhong and Jingren Zhou and Fan Zhou and Jing Zhou and Yuanzhi Zhu and Ke Zhu},
      year={2025},
      eprint={2511.21631},
      archivePrefix={arXiv},
      primaryClass={cs.CV},
      url={https://arxiv.org/abs/2511.21631}, 
}

@misc{hong2025relicinteractivevideoworld,
      title={RELIC: Interactive Video World Model with Long-Horizon Memory}, 
      author={Yicong Hong and Yiqun Mei and Chongjian Ge and Yiran Xu and Yang Zhou and Sai Bi and Yannick Hold-Geoffroy and Mike Roberts and Matthew Fisher and Eli Shechtman and Kalyan Sunkavalli and Feng Liu and Zhengqi Li and Hao Tan},
      year={2025},
      eprint={2512.04040},
      archivePrefix={arXiv},
      primaryClass={cs.CV},
      url={https://arxiv.org/abs/2512.04040}, 
}

@inproceedings{
lipman2023flow,
title={Flow Matching for Generative Modeling},
author={Yaron Lipman and Ricky T. Q. Chen and Heli Ben-Hamu and Maximilian Nickel and Matthew Le},
booktitle={ICLR},
year={2023},
}

@inproceedings{yin2025causvid,
        title={From Slow Bidirectional to Fast Autoregressive Video Diffusion Models},
        author={Yin, Tianwei and Zhang, Qiang and Zhang, Richard and Freeman, William T and Durand, Fredo and Shechtman, Eli and Huang, Xun},
        journal={CVPR},
        year={2025}
}

@article{chen2025diffusion,
  title={Diffusion forcing: Next-token prediction meets full-sequence diffusion},
  author={Chen, Boyuan and Mart{\'\i} Mons{\'o}, Diego and Du, Yilun and Simchowitz, Max and Tedrake, Russ and Sitzmann, Vincent},
  journal={NeurIPS},
  volume={37},
  pages={24081--24125},
  year={2025}
}

@article{shin2025motionstream,
  title={MotionStream: Real-Time Video Generation with Interactive Motion Controls},
  author={Shin, Joonghyuk and Li, Zhengqi and Zhang, Richard and Zhu, Jun-Yan and Park, Jaesik and Shechtman, Eli and Huang, Xun},
  journal={arXiv preprint arXiv:2511.01266},
  year={2025}
}

@inproceedings{yin2024onestep,
      title={One-step Diffusion with Distribution Matching Distillation},
      author={Yin, Tianwei and Gharbi, Micha{\"e}l and Zhang, Richard and Shechtman, Eli and Durand, Fr{\'e}do and Freeman, William T and Park, Taesung},
      booktitle={CVPR},
      year={2024}
}

@article{yang2025longlive,
      title={LongLive: Real-time Interactive Long Video Generation},
      author={Shuai Yang and Wei Huang and Ruihang Chu and Yicheng Xiao and Yuyang Zhao and Xianbang Wang and Muyang Li and Enze Xie and Yingcong Chen and Yao Lu and Song Hanand Yukang Chen},
      year={2025},
      eprint={2509.22622},
      archivePrefix={arXiv},
      primaryClass={cs.CV}
}

@misc{shah2024flashattention3fastaccurateattention,
      title={FlashAttention-3: Fast and Accurate Attention with Asynchrony and Low-precision}, 
      author={Jay Shah and Ganesh Bikshandi and Ying Zhang and Vijay Thakkar and Pradeep Ramani and Tri Dao},
      year={2024},
      eprint={2407.08608},
      archivePrefix={arXiv},
      primaryClass={cs.LG},
      url={https://arxiv.org/abs/2407.08608}, 
}

@inproceedings{loshchilov2018decoupled,
title={Decoupled Weight Decay Regularization},
author={Ilya Loshchilov and Frank Hutter},
booktitle={ICLR},
year={2019},
}

@inproceedings{barron2022mipnerf360,
    title={Mip-NeRF 360: Unbounded Anti-Aliased Neural Radiance Fields},
    author={Jonathan T. Barron and Ben Mildenhall and 
            Dor Verbin and Pratul P. Srinivasan and Peter Hedman},
    booktitle={CVPR},
    year={2022}
}

@ARTICLE{ssim,
  author={Zhou Wang and Bovik, A.C. and Sheikh, H.R. and Simoncelli, E.P.},
  journal={IEEE Transactions on Image Processing}, 
  title={Image quality assessment: from error visibility to structural similarity}, 
  year={2004},
  volume={13},
  number={4},
  pages={600-612},
  doi={10.1109/TIP.2003.819861}
}

@inproceedings{zhang2018perceptual,
  title={The Unreasonable Effectiveness of Deep Features as a Perceptual Metric},
  author={Zhang, Richard and Isola, Phillip and Efros, Alexei A and Shechtman, Eli and Wang, Oliver},
  booktitle={CVPR},
  year={2018}
}

@article{heusel2017gans,
  title={Gans trained by a two time-scale update rule converge to a local nash equilibrium},
  author={Heusel, Martin and Ramsauer, Hubert and Unterthiner, Thomas and Nessler, Bernhard and Hochreiter, Sepp},
  journal={NeurIPS},
  volume={30},
  year={2017}
}

@article{zhou2025stable,
      title={Stable Virtual Camera: Generative View Synthesis with Diffusion Models},
      author={Jensen (Jinghao) Zhou and Hang Gao and Vikram Voleti and Aaryaman Vasishta and Chun-Han Yao and Mark Boss and
      Philip Torr and Christian Rupprecht and Varun Jampani
      },
      journal={arXiv preprint},
      year={2025}
  }

@article{ho2020diffusionmodels,
    author={Jonathan Ho and Ajay Jain and Pieter Abbeel},
    title={Denoising
diffusion probabilistic models},
    journal={NeurIPS},
    year={2020} 
}

@misc{nvidia2025cosmosworldfoundationmodel,
  title     = {Cosmos World Foundation Model Platform for Physical AI},
  author    = {NVIDIA and Agarwal, Niket and Ali, Arslan and Bala, Maciej and Balaji, Yogesh and Barker, Erik and Cai, Tiffany and Chattopadhyay, Prithvijit and Chen, Yongxin and Cui, Yin and Ding, Yifan and Dworakowski, Daniel and Fan, Jiaojiao and Fenzi, Michele and Ferroni, Francesco and Fidler, Sanja and Fox, Dieter and Ge, Songwei and Ge, Yunhao and Gu, Jinwei and Gururani, Siddharth and He, Ethan and Huang, Jiahui and Huffman, Jacob and Jannaty, Pooya and Jin, Jingyi and Kim, Seung Wook and Klár, Gergely and Lam, Grace and Lan, Shiyi and Leal-Taixe, Laura and Li, Anqi and Li, Zhaoshuo and Lin, Chen-Hsuan and Lin, Tsung-Yi and Ling, Huan and Liu, Ming-Yu and Liu, Xian and Luo, Alice and Ma, Qianli and Mao, Hanzi and Mo, Kaichun and Mousavian, Arsalan and Nah, Seungjun and Niverty, Sriharsha and Page, David and Paschalidou, Despoina and Patel, Zeeshan and Pavao, Lindsey and Ramezanali, Morteza and Reda, Fitsum and Ren, Xiaowei and Sabavat, Vasanth Rao Naik and Schmerling, Ed and Shi, Stella and Stefaniak, Bartosz and Tang, Shitao and Tchapmi, Lyne and Tredak, Przemek and Tseng, Wei-Cheng and Varghese, Jibin and Wang, Hao and Wang, Haoxiang and Wang, Heng and Wang, Ting-Chun and Wei, Fangyin and Wei, Xinyue and Wu, Jay Zhangjie and Xu, Jiashu and Yang, Wei and Yen-Chen, Lin and Zeng, Xiaohui and Zeng, Yu and Zhang, Jing and Zhang, Qinsheng and Zhang, Yuxuan and Zhao, Qingqing and Zolkowski, Artur},
  journal   = {arXiv preprint arXiv:2501.03575},
  year      = {2025},
  url       = {https://arxiv.org/abs/2501.03575}
}

@misc{veo2024,
  title        = {Veo: A Generative Model for High-Quality Video},
  author       = {{Google DeepMind}},
  year         = {2024},
  howpublished = {\url{https://deepmind.google/technologies/veo/}},
  note         = {Accessed: 2025}
}

@misc{sora2024,
  title        = {Sora: Creating Video from Text},
  author       = {{OpenAI}},
  year         = {2024},
  howpublished = {\url{https://openai.com/sora}},
  note         = {Accessed: 2025}
}

@article{Knapitsch2017,
    author    = {Arno Knapitsch and Jaesik Park and Qian-Yi Zhou and Vladlen Koltun},
    title     = {Tanks and Temples: Benchmarking Large-Scale Scene Reconstruction},
    journal   = {ACM Transactions on Graphics},
    volume    = {36},
    number    = {4},
    year      = {2017},
}

@inproceedings{asim25met3r,
    title = {MEt3R: Measuring Multi-View Consistency in Generated Images},
    author = {Asim, Mohammad and Wewer, Christopher and Wimmer, Thomas and Schiele, Bernt and Lenssen, Jan Eric},
    booktitle={CVPR},
    year={2025}
  }

@inproceedings{avrahami2022blended,
    title     = {Blended Diffusion for Text-Driven Editing of Natural Images},
    author    = {Avrahami, Omri and Lischinski, Dani and Fried, Ohad},
    booktitle={CVPR},
    year={2022}
  }

@inproceedings{bahmani2026lyra,
    title     = {Lyra: Generative 3D Scene Reconstruction via Self-Distillation with Video Diffusion Models},
    author    = {Bahmani, Sherwin and Shen, Tianchang and Ren, Jiawei and Huang, Jiahui and Jiang, Yifeng and
                 Turki, Haithem and Tagliasacchi, Andrea and Lindell, David B. and Gojcic, Zan
                 and Fidler, Sanja and Ling, Huan and Gao, Jun and Ren, Xuanchi},
    booktitle = {ICLR},
    year      = {2026}
}

@inproceedings{he2025flexworld,
    title={FlexWorld: Progressively Expanding 3D Scenes for Flexiable-View Synthesis}, 
    author={Luxi Chen and Zihan Zhou and Min Zhao and Yikai Wang and Ge Zhang and Wenhao Huang and Hao Sun and Ji-Rong Wen and Chongxuan Li},
    booktitle={NeurIPS},
    year={2025}
}

@article{huang2025voyager,
  title={Voyager: Long-Range and World-Consistent Video Diffusion for Explorable 3D Scene Generation},
  author={Wan, Haoran and Zhang, Jian and Zhang, Richard and Luo, Jia-Bin and Fang, Xinyang and Yang, Lingbo and Cao, Yanpei and Shan, Ying},
  journal={ACM Transactions on Graphics},
  year={2025},
}

@inproceedings{kim2025rad,
    title     = {RAD: Region-Aware Diffusion Models for Image Inpainting},
    author    = {Kim, Sora and Suh, Sungho and Lee, Minsik},
    booktitle={CVPR},
    year={2025}
  }

@inproceedings{leroy2024mast3r,
    author = {Leroy, Vincent and Cabon, Yohann and Revaud, Jerome},
    title = {Grounding Image Matching in 3D with MASt3R},
    booktitle = {ECCV},
    year = {2024}
}

@inproceedings{li2024dngaussian,
    title={DNGaussian: Optimizing Sparse-View 3D Gaussian Radiance Fields with Global-Local Depth Normalization}, 
    author={Jiahe Li and Jiawei Zhang and Xiao Bai and Jin Zheng and Xin Ning and Jun Zhou and Lin Gu},
    booktitle={CVPR},
    year={2024}
}

@inproceedings{li2025vmem,
  title={VMem: Consistent Interactive Video Scene Generation with Surfel-Indexed View Memory},
  author={Li, Runjia and Torr, Philip and Vedaldi, Andrea and Jakab, Tomas},
  booktitle = {ICCV},
  year={2025}
}

@article{liu2024reconx,
    title={{ReconX}: Reconstruct Any Scene from Sparse Views with Video Diffusion Model},
    author={Liu, Fangfu and Wu, Wenqiang and Tan, Hanyang and Yuan, Yueqi and Zhou, Yikai and Liu, Junwei and Duan, Kangjie and Xie, Haowen and Pei, Jingwen and Wang, He and others},
    journal={IEEE Transactions on Image Processing},
    year={2026}
}

@article{lu2025matrix3d,
  title={Matrix3D: Large Photogrammetry Model All-in-One},
  author={Lu, Yuanxun and Zhang, Jingyang and Fang, Tian and Nahmias, Jean-Daniel and Tsin, Yanghai and Quan, Long and Cao, Xun and Yao, Yao and Li, Shiwei},
  journal={CVPR},
  year={2025}
}

@inproceedings{mayet2025tdpaint,
    title={TD-Paint: Faster Diffusion Inpainting Through Time Aware Pixel Conditioning}, 
    author={Tsiry Mayet and Pourya Shamsolmoali and Simon Bernard and Eric Granger and Romain Hérault and Clement Chatelain},
    booktitle={ICLR},
    year={2025}
  }

@inproceedings{teed20202raft,
    author = {Teed, Zachary and Deng, Jia},
    title = {RAFT: Recurrent All-Pairs Field Transforms for Optical Flow},
    booktitle = {ECCV},
    year = {2020}
}

@inproceedings{wen2023treering,
title={Tree-Ring Watermarks: Fingerprints for Diffusion Images that are Invisible and Robust}, 
author={Yuxin Wen and John Kirchenbauer and Jonas Geiping and Tom Goldstein},
booktitle={NeurIPS},
year={2023}
}

@inproceedings{wu2025spmem,
    title={Video world models with long-term spatial memory},
    author={Wu, Tong and Yang, Shuai and Po, Ryan and Xu, Yinghao and Liu, Ziwei and Lin, Dahua and Wetzstein, Gordon},
    booktitle={NeurIPS},
    year={2025}
}

@inproceedings{zhai2025stargen,
    title={StarGen: A Spatiotemporal Autoregression Framework with Video Diffusion Model for Scalable and Controllable Scene Generation}, 
    author={Shangjin Zhai and Zhichao Ye and Jialin Liu and Weijian Xie and Jiaqi Hu and Zhen Peng and Hua Xue and Danpeng Chen and Xiaomeng Wang and Lei Yang and Nan Wang and Haomin Liu and Guofeng Zhang},
    booktitle={CVPR},
    year={2025}
  }

@inproceedings{zhuang2026flashvsr,
  title={FlashVSR: Towards Real-Time Diffusion-Based Streaming Video Super-Resolution},
  author={Zhuang, Junhao and Guo, Shi and Cai, Xin and Li, Xiaohui and Liu, Yihao and Yuan, Chun and Xue, Tianfan},
  booktitle={CVPR},
  year={2026}
}

\clearpage
\appendix
\twocolumn[\section*{Supplementary Material}\vspace{1em}]
\section{Opacity Mixing and Flow Matching}
\label{sec:opacity-mixing-derivation}

Our opacity mixing strategy is fully compatible with the conditional flow matching (CFM) framework~\cite{lipman2023flowmatching} as the CFM loss $\mathbb{E}_{t,\mathbf{z}_0,\mathbf{z}_1}\bigl\lVert \mathbf{v}_\theta(\mathbf{z}_t, t, \text{cond}) - (\mathbf{z}_1 - \mathbf{z}_0)\bigr\rVert^2$ is valid for \emph{any} joint distribution $q(\mathbf{z}_0, \mathbf{z}_1)$, not only $\mathbf{z}_0 \sim \mathcal{N}(\mathbf{0}, \mathbf{I})$. In our setting, we define the source sample as:
  \begin{equation}
  \mathbf{z}_0 \coloneqq \mathbf{O}_z \mathbf{z}_{deg} + (1 - \mathbf{O}_z) \boldsymbol{\epsilon}, \quad \boldsymbol{\epsilon} \sim \mathcal{N}(\mathbf{0}, \mathbf{I}),
  \end{equation}
  where $\mathbf{O}_z$ is the spatially varying, downscaled opacity map and $\mathbf{z}_{deg}$ is the VAE-encoded degraded rendering. Let $\mathbf{z}_1$ denote the clean target latent. We sample a global scalar $t \sim \mathcal{U}[0,1]$
   and form the interpolant:
  \begin{equation}
  \mathbf{z}_t = (1-t)\,\mathbf{z}_0 + t\,\mathbf{z}_1,
  \end{equation}
  with target velocity $\mathbf{v}_t = \mathbf{z}_1 - \mathbf{z}_0$. The spatial variation introduced by $\mathbf{O}_z$ is encoded entirely in $\mathbf{z}_0$ and consequently propagates to both $\mathbf{z}_t$ and the target velocity $\mathbf{v}_t$, not to the scalar time variable $t$. No per-location timestep conditioning is required: the network receives $(\mathbf{z}_t, t, \text{cond})$ with a single global $t$, exactly as in standard flow matching.

  At inference, we draw $\mathbf{z}_0 \sim q(\mathbf{z}_0)$ using the same opacity mixing procedure and integrate the learned ODE from $t=0$ to $t=1$ using the same global time parameterization.

\section{Text Conditioning}
\label{sec:conditioning}

  \begin{table}
    \centering
    
    \resizebox{0.8\linewidth}{!}{
\begin{tabular}{@{}l|rrr@{}}
      \toprule
      Dataset & $\Delta$PSNR & $\Delta$SSIM & $\Delta$LPIPS \\
      \midrule
      Mip-NeRF 360 (3 views) & +0.14 & +0.003 & $-$0.002 \\
      Mip-NeRF 360 (6 views) & +0.07 & +0.002 & $-$0.001 \\
      Mip-NeRF 360 (9 views) & +0.03 & +0.003 & $-$0.001 \\
      \hline
      DL3DV & +0.02 & 0.000 & $-$0.001 \\
      Nerfbusters & $-$0.07 & +0.001 & 0.000 \\
      \bottomrule
    \end{tabular}
    }
    \caption{\textbf{Text conditioning.} We measure the impact of VLM-generated prompts vs.\ no prompt for \methodTDp. Text prompts provide a small benefit in sparse settings that diminishes with denser captures.}
    \label{tab:text-conditioning}
    
  \end{table}

We further quantify the contribution of text conditioning by comparing \methodTDp results with and without VLM-generated prompts in \cref{tab:text-conditioning}. Text conditioning provides a minor benefit in the most sparse settings (+0.14\,dB PSNR on Mip-NeRF 360 with 3 views), but this effect diminishes with denser captures.

\begin{table}
  \centering
  
      \resizebox{0.9\linewidth}{!}{
      \begin{tabular}{@{}l|c|cccc@{}}
        \toprule
        & & \multicolumn{4}{c}{FPS $\uparrow$} \\
        Method & GPUs & 1 step & 2 steps & 3 steps & 4 steps \\
        \midrule
        \method (14B) & 1 & 29.42 & 16.07 & 11.03 & 8.36 \\
        \method (14B) & 4 & 58.72 & 35.91 & 24.65 & 19.18 \\
        \method (1.3B) & 1 & 86.75 & 57.76 & 43.20 & 34.38 \\
        \method (1.3B) & 4 & 101.77 & 69.44 & 53.77 & 49.24 \\
        \bottomrule
      \end{tabular}
      }
    \caption{\textbf{Inference configurations.} Fewer denoising steps and context parallelism across multiple GPUs further improve throughput, with the 1.3B variant reaching up to 101.77 FPS.}
    \label{tab:timing-full}
  
  \end{table}

\section{Denoising Steps}
\label{sec:denoising-steps}

\begin{figure*}
  \centering
    \includegraphics[width=0.8\textwidth]{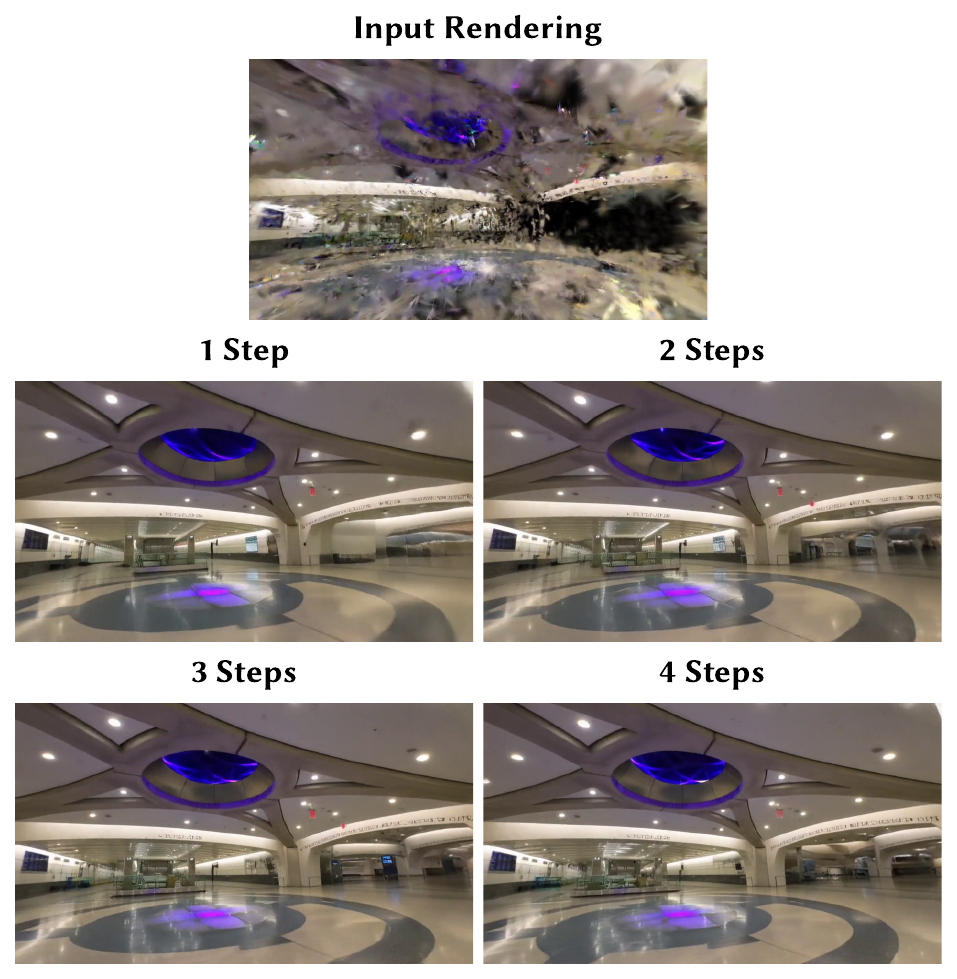}
\caption{\textbf{Denoising steps}. We vary the number of denoising steps when beginning from the initial degraded rendering. \method\ can render plausible content in as few as 1 step, although sharpness and temporal consistency suffer somewhat in empty areas.}
    \label{fig:artifixer-denoising-steps}
\end{figure*}

As \method starts from renderings instead of pure noise, it is able to generate plausible visuals in fewer than four steps in most cases. Reducing the number of denoising steps significantly improves throughput, with context parallelism across multiple GPUs providing further gains (\cref{tab:timing-full}). However, sharpness and temporal consistency suffer somewhat in empty areas (\cref{fig:artifixer-denoising-steps}). This is largely mitigated when revisiting previously explored areas in our \methodTD and \methodTDp variants, as the 3D distillation process provides strong conditioning for subsequent generations.

\begin{table*}
    \centering
    
    \resizebox{0.8\linewidth}{!}{
        \begin{tabular}{@{}l@{\,\,}|ccc|ccc|ccc@{}}
      \toprule
      & \multicolumn{3}{c|}{PSNR $\uparrow$} & \multicolumn{3}{c|}{SSIM $\uparrow$} & \multicolumn{3}{c}{LPIPS $\downarrow$} \\
      Method & 3-view & 6-view & 9-view & 3-view & 6-view & 9-view & 3-view & 6-view & 9-view \\
      \midrule
      GenFusion~\cite{Wu2025GenFusion} & 15.29 & 17.16 & 18.36 & 0.369 & \cellcolor{yellow}0.447 & \cellcolor{yellow}0.496 & 0.585 & 0.500 & 0.465 \\
      GSFixer~\cite{yin2025gsfixerimproving3dgaussian} & 15.61 & 17.27 & 18.63 & 0.370 & 0.426 & 0.481 & 0.559 & \cellcolor{yellow}0.478 & \cellcolor{yellow}0.420 \\
      CAT3D~\cite{gao2024cat3d} & \cellcolor{orange}16.62 & \cellcolor{yellow}17.72 & \cellcolor{yellow}18.67 & \cellcolor{yellow}0.377 & 0.425 & 0.460 & \cellcolor{yellow}0.515 & 0.482 & 0.460 \\
      \hline
      \textbf{\methodTDp (1.3B)} & \cellcolor{yellow}16.60 & \cellcolor{orange}18.04 & \cellcolor{orange}19.44 & \cellcolor{orange}0.414 & \cellcolor{orange}0.466 & \cellcolor{orange}0.513 & \cellcolor{orange}0.486 & \cellcolor{orange}0.435 &
  \cellcolor{orange}0.394 \\
      \textbf{\methodTDp (14B)} & \cellcolor{red}17.51 & \cellcolor{red}18.95 & \cellcolor{red}20.16 & \cellcolor{red}0.444 & \cellcolor{red}0.498 & \cellcolor{red}0.537 & \cellcolor{red}0.441 & \cellcolor{red}0.396 & \cellcolor{red}0.359 \\
      \bottomrule
    \end{tabular}
    }
    \caption{\textbf{Impact of model scale on Mip-NeRF 360.} Our 1.3B variant matches CAT3D within 0.02\,dB on the 3-view split and exceeds other video model baselines despite using fewer parameters.}
    \label{tab:scale-m360}
    
  \end{table*}

  \begin{table}
    \centering
    
    \resizebox{0.8\linewidth}{!}{
    \begin{tabular}{@{}l|cccc@{}}
      \toprule
      Method & PSNR $\uparrow$ & SSIM $\uparrow$ & LPIPS $\downarrow$ & FID $\downarrow$ \\
      \midrule
      GenFusion~\cite{Wu2025GenFusion} & \cellcolor{yellow}17.03 & \cellcolor{yellow}0.624 & \cellcolor{yellow}0.392 & 132.91 \\
      Gen3C~\cite{ren2025gen3c} & 15.50 & 0.491 & 0.476 & \cellcolor{yellow}68.36 \\
      \hline
      \textbf{\methodTDp (1.3B)} & \cellcolor{orange}19.04 & \cellcolor{orange}0.635 & \cellcolor{orange}0.352 & \cellcolor{orange}22.3 \\
      \textbf{\methodTDp (14B)} & \cellcolor{red}20.15 & \cellcolor{red}0.662 & \cellcolor{red}0.307 & \cellcolor{red}13.91 \\
      \bottomrule
    \end{tabular}
    }
    \caption{\textbf{Impact of model scale on novel content generation (DL3DV).} Even with a 1.3B backbone, \methodTDp outperforms the other video model baselines by a wide margin.}
    \label{tab:scale-dl3dv}
    
  \end{table}

  \begin{table}
    \centering

    \resizebox{0.8\linewidth}{!}{
    \begin{tabular}{@{}l|ccc@{}}
      \toprule
      Method & PSNR $\uparrow$ & SSIM $\uparrow$ & LPIPS $\downarrow$ \\
      \midrule
      3DGS~\cite{kerbl3Dgaussians} & 9.57 & 0.108 & 0.779 \\
      SparseNeRF~\cite{wang2023sparsenerf} & 9.23 & \cellcolor{yellow}0.191 & \cellcolor{yellow}0.632 \\
      DNGaussian~\cite{li2024dngaussian} & \cellcolor{yellow}10.23 & 0.156 & 0.643 \\
      ReconX~\cite{liu2024reconx} & \cellcolor{orange}14.28 & \cellcolor{orange}0.394 & \cellcolor{orange}0.564 \\      \hline
      \textbf{\methodTDp} & \cellcolor{red}14.75 & \cellcolor{red}0.464 & \cellcolor{red}0.463 \\
      \bottomrule
    \end{tabular}
    }
    \caption{\textbf{Tanks and Temples (2-view).} \methodTDp outperforms all baselines.}
    \label{tab:tanks-and-temples}

  \end{table}

  \begin{table}
    \centering
    
    \resizebox{0.75\linewidth}{!}{
    \begin{tabular}{@{}l|cc@{}}
      \toprule
      Method & MASt3R $\downarrow$ & RAFT $\downarrow$ \\
      \midrule
      Fixer~\cite{nvidia-fixer} & 0.1288 & 0.1236 \\
      Difix3D+~\cite{wu2025difix3d+} & 0.0974 & 0.0959 \\
      GenFusion~\cite{Wu2025GenFusion} & 0.0817 & 0.0786 \\
      Gen3C~\cite{ren2025gen3c} & 0.0766 & 0.0757 \\
      \hline
    \textbf{\method} & \cellcolor{yellow}0.0749 & \cellcolor{yellow}0.0749 \\
    \textbf{\methodTDp} & \cellcolor{orange}0.0697 & \cellcolor{orange}0.0697 \\
    \textbf{\methodTD} & \cellcolor{red}0.0646 & \cellcolor{red}0.0647 \\
      \bottomrule
    \end{tabular}
    }
    \caption{\textbf{Multi-view consistency.} We measure multi-view consistency via MEt3R~\cite{asim25met3r} with MASt3R and RAFT backbones. All \method variants outperform baselines, with \methodTD achieving the best results due to its explicit multi-view-consistent 3D representation.}
    \label{tab:met3r}
    
  \end{table}

\section{Additional Experiments}
\label{sec:additional-experiments}

\paragraph{Model scale.} To disentangle the contribution of our method from backbone capacity, we train the full pipeline with Wan 2.1 T2V-1.3B and report \methodTDp results in \cref{tab:scale-m360,tab:scale-dl3dv}. For reference, GenFusion~\cite{Wu2025GenFusion} uses a 1.4B-parameter backbone, GSFixer~\cite{yin2025gsfixerimproving3dgaussian} 5B, and Gen3C~\cite{ren2025gen3c} 7B. Our 1.3B variant matches CAT3D~\cite{gao2024cat3d} within 0.02\,dB on the 3-view Mip-NeRF 360 split and exceeds all other baselines.

\paragraph{Tanks and Temples.} To further evaluate generalization, we report results on the Tanks and Temples dataset~\cite{Knapitsch2017} using the 2-view setting from ReconX~\cite{liu2024reconx} in \cref{tab:tanks-and-temples}.

\paragraph{Multi-view consistency.} We evaluate multi-view consistency using MEt3R~\cite{asim25met3r} with MASt3R~\cite{leroy2024mast3r} depth-based reprojection and RAFT~\cite{teed20202raft} optical flow-based warping backbones in \cref{tab:met3r}. All \method variants outperform baselines, with \methodTD achieving the best consistency due to its explicit 3D representation.

\section{Limitations}
While \method reaches interactive rates, it remains significantly slower than direct rendering from neural scene representations. Decoding in temporal chunks also introduces latency that may be undesirable for applications such as embodied AI. Additionally, the \method and \methodTDp variants are limited to 720p by the backbone video model, whereas \methodTD renders at the native resolution of the underlying 3D representation. As with other video diffusion models, our method can occasionally blur fine details and text, and may introduce subtle color shifts when the rendering condition is absent or highly degraded. Promising directions for future work include further reducing denoising steps, enabling single-frame decoding while maintaining temporal coherence, and applying video super-resolution~\cite{zhuang2026flashvsr} to close the resolution gap.

\section{Societal Impact}
\method synthesizes photorealistic scene content and can plausibly inpaint unobserved regions, raising concerns about potential misuse for generating deceptive visual media. Appropriate safeguards such as watermarking generated content~\cite{wen2023treering} should be considered for deployment. From an environmental perspective, training our 14B-parameter model requires approximately 15k GPU-hours on H100 hardware. Our truncated training schedule achieves near-full quality at roughly 25\% of this cost, and our 1.3B-parameter variant further reduces training compute while remaining competitive with prior work.

\section{Sparse Reconstruction}
\label{sec:sparse-reconstruction}

\paragraph{Camera Sampling.} We describe our camera sampling strategy in \cref{alg:camera_sampling}. Given a set of camera poses $\mathbf{P}$, we define the pairwise distance between two poses as $d = ||\mathbf{R}_i - \mathbf{R}_j||_{F} +||\mathbf{t}_i - \mathbf{t}_j||_2$. We initialize the clustering process by identifying the pair $(P_1, P_2)$ that maximizes this distance and using them as seeds for groups $G_1$ and $G_2$. The remaining cameras are assigned to the group of their nearest seed. Finally, to evaluate varying levels of sparsity, we apply farthest point sampling within each group to select subsets of size $K = \{2, \cdots, 12\}$.

\noindent
\begin{algorithm}[h]
\SetKwInput{KwInput}{Input}
\SetKwInput{KwOutput}{Output}
\SetKwComment{Comment}{// }{}

\caption{CameraSampling}
\label{alg:camera_sampling}

\KwInput{Camera poses $\mathbf{P}$, Selection count $K$, Distance function $d$}
\KwOutput{Selected subsets $\mathcal{S}_1 \subset G_1$ and $\mathcal{S}_2 \subset G_2$}

\tcc{1. Find global farthest camera pair}
$(P_1, P_2) \gets \operatorname*{argmax}_{P_i, P_j \in \mathbf{P}} d(P_i, P_j)$\;

\tcc{2. Cluster: Assign cameras to nearest seed camera}
$G_1 \gets \{ P \in \mathbf{P} \mid D(P, P_1) \le D(P, P_2) \}$\;
$G_2 \gets \mathbf{P} \setminus G_1$\;

\tcc{3. Select Top-K points in EACH group}
\ForEach{$i \in \{1, 2\}$}{
    $\mathcal{S}_i \gets \{ P_i \}$ \Comment*[r]{Start with the seed camera}
    \While{$|\mathcal{S}_i| < K$ and $|\mathcal{S}_i| < | G_i|$}{
        \tcc{Find pose maximizing distance to current selection}
        $P_{next} \gets \operatorname*{argmax}_{P \in G_i \setminus \mathcal{S}_i} \left( \min_{s \in \mathcal{S}_i} D(P, s) \right)$\;
        $\mathcal{S}_i \gets \mathcal{S}_i \cup \{ P_{next} \}$\;
    }
}

\Return{$\mathcal{S}_1, \mathcal{S}_2$}

\end{algorithm}

\paragraph{Reconstruction.} We generate the initial reconstructions we pass to the \method model using the official 3DGUT implementation~\cite{wu20253dgut} with MCMC~\cite{kheradmand20243d} sampling (reconstructions used during training are prepared offline). We run each reconstruction for 10,000 iterations, taking slightly less than 10 minutes per reconstruction.

\paragraph{Captioning.} We generate captions for each DL3DV scene from Qwen3-VL-30B-A3B-Instruct~\cite{bai2025qwen3vltechnicalreport} on different frame subsets to encourage prompt diversity. Similar to \cite{hong2025relicinteractivevideoworld}, we suppress descriptions of ego-camera movement to avoid entanglement with camera ray conditioning. We use the prompt below:

\begin{small}
\hfill \break

    You are a video captioning specialist whose goal is to generate high-quality English prompts by referring to the details of the user's input videos. Your task is to carefully analyze the content, context, and actions within the video, and produce a complete, expressive, and natural-sounding caption that accurately conveys the scene. The caption should preserve the original intent and meaning of the video while enhancing its clarity and descriptive richness. Strictly adhere to the formatting of the examples provided.

    Task Requirements:
    1. You need to describe the main subject of the video in detail, including their appearance, actions, expressions, and the surrounding environment.
    2. You should never describe any details about the camera movement or camera angles.
    3. Your output should convey natural movement attributes, incorporating natural actions related to the described subject category, using simple and direct verbs as much as possible.
    4. You should reference the detailed information in the video, such as character actions, clothing, backgrounds, and emphasize the details in the photo.
    5. Control the output prompt to around 80-100 words.
    6. No matter what language the user inputs, you must always output in English.

    Example of the English prompt:
    1. A Japanese fresh film-style photo of a young East Asian girl with double braids sitting by the boat. The girl wears a white square collar puff sleeve dress, decorated with pleats and buttons. She has fair skin, delicate features, and slightly melancholic eyes, staring directly at the camera. Her hair falls naturally, with bangs covering part of her forehead. She rests her hands on the boat, appearing natural and relaxed. The background features a blurred outdoor scene, with hints of blue sky, mountains, and some dry plants. The photo has a vintage film texture. A medium shot of a seated portrait.
    2. An anime illustration in vibrant thick painting style of a white girl with cat ears holding a folder, showing a slightly dissatisfied expression. She has long dark purple hair and red eyes, wearing a dark gray skirt and a light gray top with a white waist tie and a name tag in bold Chinese characters. The background has a light yellow indoor tone, with faint outlines of some furniture visible. A pink halo hovers above her head, in a smooth Japanese cel-shading style. A close-up shot from a slightly elevated perspective.
    3. CG game concept digital art featuring a huge crocodile with its mouth wide open, with trees and thorns growing on its back. The crocodile's skin is rough and grayish-white, resembling stone or wood texture. Its back is lush with trees, shrubs, and thorny protrusions. With its mouth agape, the crocodile reveals a pink tongue and sharp teeth. The background features a dusk sky with some distant trees, giving the overall scene a dark and cold atmosphere. A close-up from a low angle.
    4. In the style of an American drama promotional poster, Walter White sits in a metal folding chair wearing a yellow protective suit, with the words "Breaking Bad" written in sans-serif English above him, surrounded by piles of dollar bills and blue plastic storage boxes. He wears glasses, staring forward, dressed in a yellow jumpsuit, with his hands resting on his knees, exuding a calm and confident demeanor. The background shows an abandoned, dim factory with light filtering through the windows. There's a noticeable grainy texture. A medium shot with a straight-on close-up of the character.

    Directly output the English text.
\end{small}

\end{document}